\documentclass[11pt]{article}

\usepackage[preprint]{acl}

\usepackage{times}
\usepackage{latexsym}
\usepackage{amsmath}
\usepackage{amsfonts}
\usepackage{amssymb}
\usepackage{subfigure}
\usepackage{xcolor}
\usepackage[dvipsnames]{xcolor}
\usepackage[skins,breakable]{tcolorbox}
\usepackage{longtable}
\usepackage{cuted}
\usepackage{enumitem}
\usepackage[normalem]{ulem}
\tcbuselibrary{breakable}

\definecolor{PromptBlue}{RGB}{85, 150, 210}    
\definecolor{PromptGreen}{RGB}{110, 170, 120}  
\definecolor{PromptOrange}{RGB}{215, 150, 90}  
\definecolor{PromptPurple}{RGB}{150, 120, 185}  
\definecolor{PromptTeal}{RGB}{90, 160, 165}     

\usepackage{soul}
\sethlcolor{Apricot!45}
\usepackage[hang,flushmargin]{footmisc}
\usepackage[T1]{fontenc}

\usepackage[utf8]{inputenc}

\usepackage{microtype}

\usepackage{inconsolata}

\usepackage{graphicx}
\usepackage{todonotes}

\usepackage{verbatim}
\usepackage{enumitem}
\setlist[itemize]{topsep=2pt, itemsep=2pt, parsep=0pt, partopsep=0pt}
\usepackage{float}
\title{MOSAIC: Modular Opinion Summarization using Aspect Identification and Clustering}

\author{Piyush Kumar Singh \\
  Viator, TripAdvisor / London \\
  \texttt{pkumarsingh@tripadvisor.com} \\\And
  Jayesh Choudhari \\
  Viator, TripAdvisor / London \\
  \texttt{jchoudhari@tripadvisor.com} \\}

\begin{document}
\maketitle
\begin{abstract}
Reviews are central to how travelers evaluate products on online marketplaces, yet existing summarization research often emphasizes end-to-end quality while overlooking benchmark reliability and the practical utility of granular insights.
To address this, we propose MOSAIC, a scalable, modular framework designed for industrial deployment that decomposes summarization into interpretable components, including theme discovery, structured opinion extraction, and grounded summary generation.
We validate the practical impact of our approach through online A/B tests on live product pages, showing that surfacing intermediate outputs improves customer experience and delivers measurable value even prior to full summarization deployment.
We further conduct extensive offline experiments to demonstrate that MOSAIC achieves superior aspect coverage and faithfulness compared to strong baselines for summarization. 
Crucially, we introduce opinion clustering as a system-level component and show that it significantly enhances faithfulness, particularly under the noisy and redundant conditions typical of user reviews.
Finally, we identify reliability limitations in the standard SPACE dataset and release a new open-source tour experience dataset (TRECS) to enable more robust evaluation. 
\begin{figure}[ht]
    \centering
    \begin{subfigure}
        \centering
        \includegraphics[width=\linewidth]{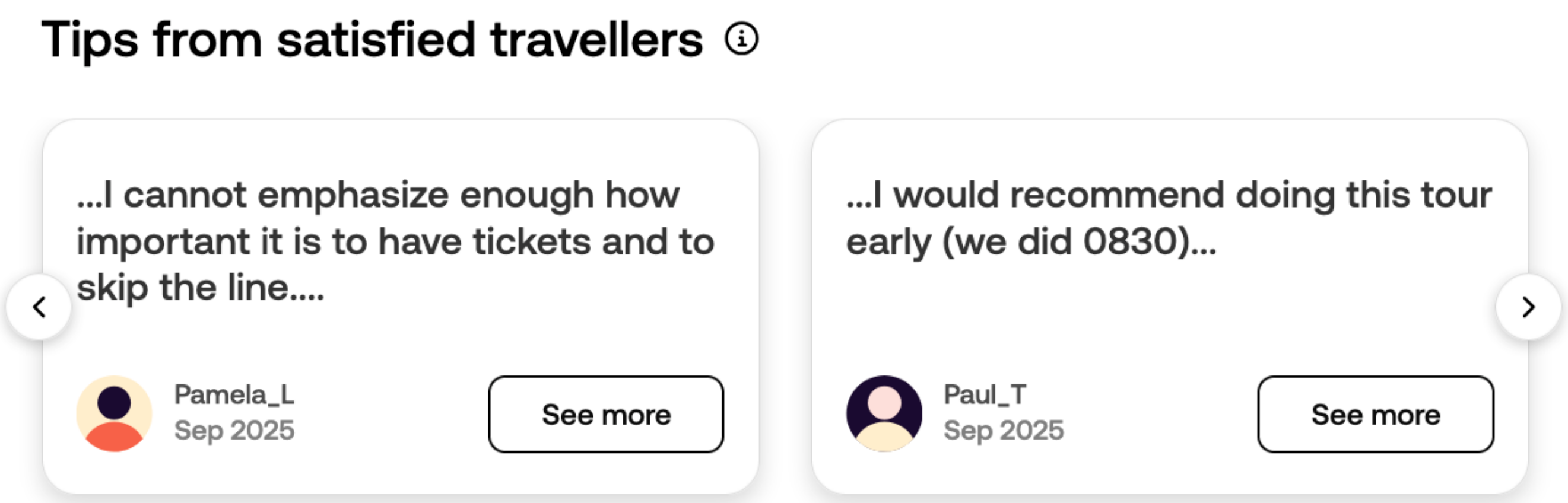}
        \label{fig:tips}
    \end{subfigure}
    \hfill
    \begin{subfigure}
        \centering
        \includegraphics[width=\linewidth]{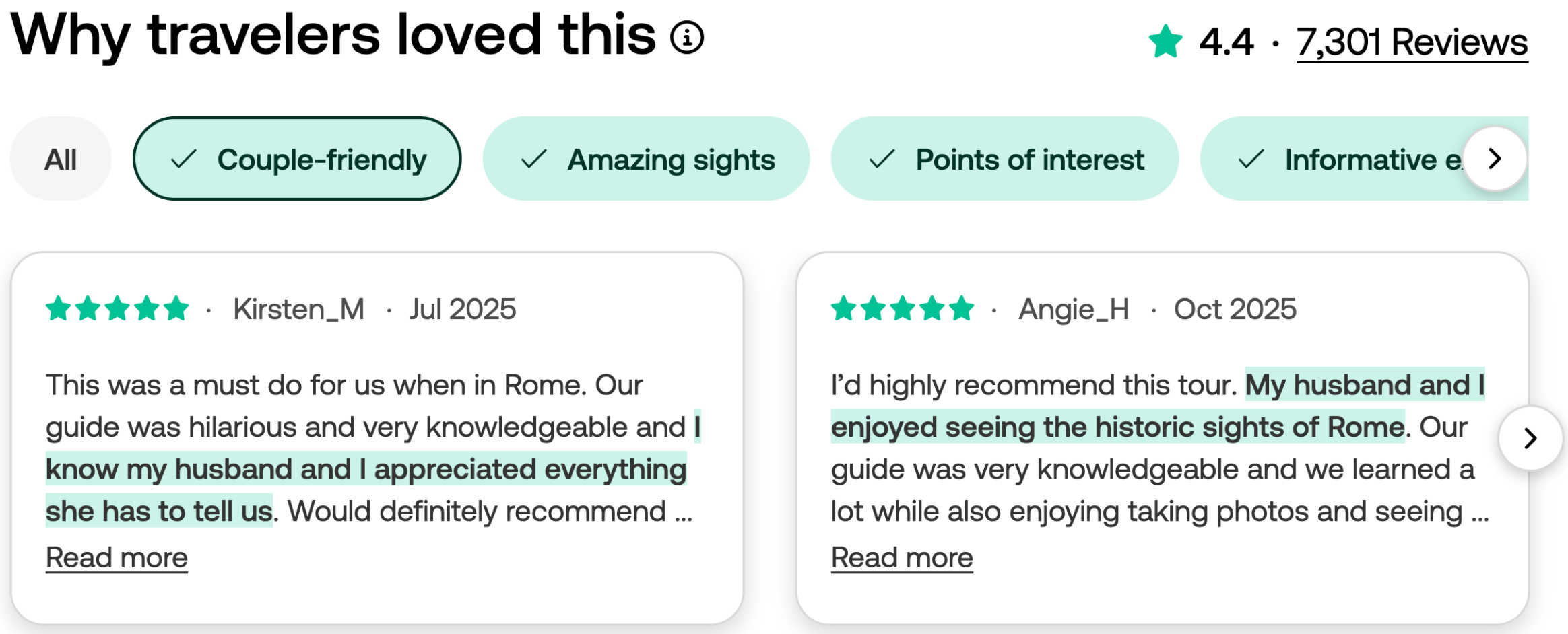}
        \label{fig:shelf}
    \end{subfigure}
    \caption{Industrial deployment of MOSAIC on a live product page. The interface surfaces intermediate outputs directly to users, including actionable Traveler Tips and interactive Review Themes (e.g., Couple-friendly, Amazing Sights) that highlight relevant text using opinions.} 
    \label{fig:industrial_application}
\end{figure}
\end{abstract}

\section{Introduction}


Online marketplaces host vast volumes of user-generated reviews that shape how customers discover, evaluate, and select products. Beyond influencing purchases, reviews provide large-scale feedback on product quality and operational performance, yet they are often underutilized for guiding product presentation and improvement. As competition grows, systematically surfacing review-driven insights is critical for user trust and marketplace performance.

Existing research has largely focused on extracting aspects from reviews or generating aggregated summaries. While early work showed language models can perform such tasks via zero-shot generation \cite{radford2019language}, recent advances in large language models enable long-context opinion summarization through direct prompting \cite{touvron2023llama}. However, evaluation often remains limited to offline metrics, with little focus on how intermediate outputs such as aspects, sentiments, or opinions impact customer-facing product pages. This creates a gap between technical capability and product impact. Effective use of reviews on product pages requires modular, transparent, and granular representations, but practical evaluation of review-driven insights remains underexplored.

We address this gap with an end-to-end, modular framework treating review understanding as a sequence of interpretable, measurable components. Unlike prior work focusing on final summaries in isolation \cite{li2025decomposed, zhou2025aspect}, our approach surfaces intermediate outputs to users (See Figure~\ref{fig:industrial_application}). Evaluations, both offline and in real product settings, show that granular, transparent review insights improve customer evaluation.




\noindent Our contributions are as follows:
\begin{itemize}[leftmargin=*]
    \item We introduce MOSAIC, a scalable and modular review summarization framework for industrial deployment, and demonstrate that it matches or outperforms state-of-the-art methods on public benchmarks.
	\item We validate intermediate outputs from MOSAIC in production through online A/B tests on live product pages.
    \item Motivated by limitations of existing benchmarks such as SPACE, we release TRECS (\textbf{T}our-experiences \textbf{RE}views \textbf{C}orpus for \textbf{S}ummarization), an open-source dataset of 344 tour products and ~140K reviews, with annotated themes, human-validated summaries, and full codebase. [\href{https://anonymous.4open.science/r/mosaic-review-summarization-78C0/}{Code Repo} ] 
    \item We introduce opinion clustering as a system-level component and demonstrate its impact on improving summary faithfulness. 
\end{itemize}

\section{Related Work}

Text summarization is broadly categorized into extractive methods, which select salient text spans from source documents \cite{mihalcea2004textrank, rossiello2017centroid, belwal2021new, di2014hybrid}, and abstractive methods, which synthesize information across inputs into coherent summaries \cite{isonuma2019unsupervised, chu2019meansum, coavoux2019unsupervised}. Opinion summarization \cite{hu2006opinion}, \cite{wang2016neural} builds on these approaches, by focusing on creating summaries based on opinions that are popular across reviews \cite{angelidis2018summarizing}, \cite{amplayo2020unsupervised}, \cite{bravzinskas2020unsupervised}, \cite{amplayo2021aspect}, \cite{tian2019aspect}, \cite{suhara2020opiniondigest}, but the lack of interpretable intermediate representations often makes verifying faithfulness difficult.
To manage this complexity, recent work has drawn motivation from chain-of-thought prompting and task decomposition \cite{wei2022chain}, \cite{khot2022decomposed}, \cite{zhou2022least} which show that breaking complex reasoning problems into simpler sub-tasks improves performance. In the context of reviews, \cite{zeng2024scientific} demonstrate that decomposing the summarization task enhances faithfulness and reduces hallucinations. Along similar lines, \cite{li2025decomposed} provided empirical evidence that the intermediate reasoning steps like opinion extraction and consolidation are crucial for assisting humans in processing large volumes of reviews. Similarly, in the specific domain of scientific reviews, \cite{li2024sentiment} propose a three-layer sentiment consolidation framework that models how meta-reviewers aggregate facet-level judgments (e.g., novelty, soundness) into meta-reviews. 

Our work is most comparable to \cite{li2025decomposed} and \cite{zhou2025aspect}. \cite{li2025decomposed} propose a robust opinion summarization framework but do not address large-scale theme extraction or issues related to opinion redundancy, both of which are central to MOSAIC. \cite{zhou2025aspect} propose a domain-agnostic approach for extracting aspects, sentiments, and supporting evidence, followed by clustering similar evidence spans. However, their summaries are constructed by concatenating evidence sentences from different arguments, which may lack smooth narrative flow, and their work does not include a comparison against other state-of-the-art approaches.
Finally, our theme extraction component mirrors the approach of \cite{tang2024prompted} which proposed a fully automated key-point extraction approach using few-shot prompting but unlike themes, key points capture finer-grained statements and are not used to construct either theme-level or product-level summaries in their work.

\section{Methodology}
We decompose the task into three modules, namely Theme Discovery \& Standardization, Theme-Constrained Opinion Extraction, and Opinion-Aware Review Summarization.

\subsection{Theme Discovery \& Standardization}
\label{subsec:theme_discover_standardisation}

\subsubsection{Unconstrained Theme Generation}
\label{subsubsec:unconstrained_theme_gen}

\begin{figure}[ht]
    \centering
\includegraphics[width=1\linewidth]{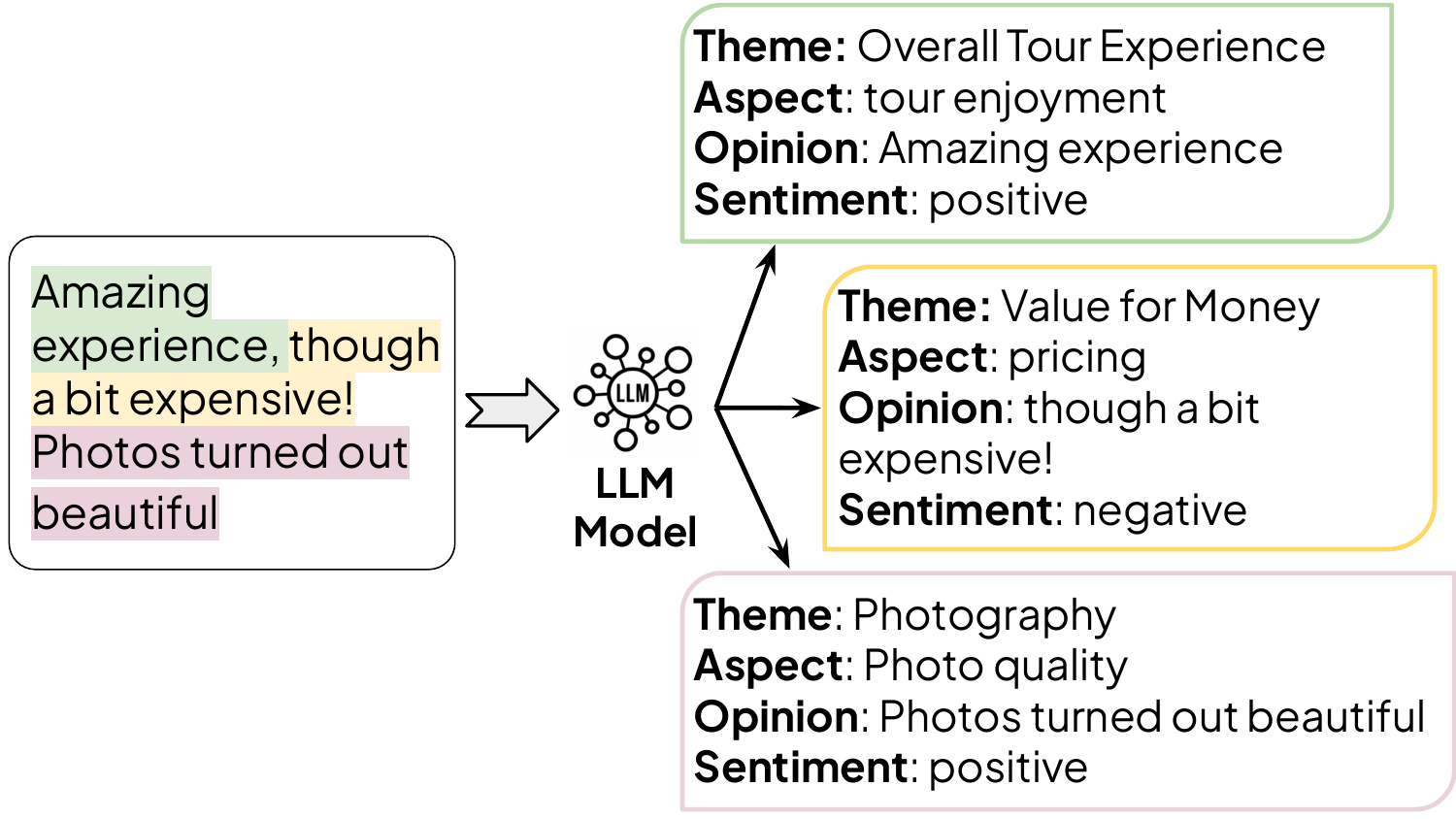}
    \caption{Example of structured ABSA output, where the LLM decomposes a single review into multiple theme–aspect–opinion–sentiment tuples that capture distinct facets of the customer experience.} 
    \label{fig:absa_output}
\end{figure}

Different domains exhibit distinct, user-relevant themes. To capture this diversity, we first employ a few-shot LLM approach to infer themes directly from the review text. The model produces structured JSON outputs aligned with the Aspect-Based Sentiment Analysis (ABSA) framework~\cite{varia2022instruction}, extracting tuples of the form (theme, aspect, opinion, sentiment) (see Figure~\ref{fig:absa_output}). This step ensures that we capture the full breadth and diversity of themes expressed across the reviews. We used GPT-4o-mini for this step across all datasets.

\subsubsection{Theme Refinement and Consolidation}
\label{subsubsec:theme_refinement}
To ensure scalability without compromising quality, we design a systematic pipeline that analyzes all extracted themes in section~\ref{subsubsec:unconstrained_theme_gen} and identifies the most important ones based on both frequency and user relevance. We construct a high-quality theme set through a three-stage filtering process:
\begin{itemize}[leftmargin=*]
\item \textbf{Frequency-Based Filtering:}
The initial set of
themes obtained in section \ref{subsubsec:unconstrained_theme_gen} is filtered based on oc-
currence frequency, removing rare or overly spe-
cific themes that contribute limited value. This
yields a consolidated set of high-frequency, user-
relevant themes that balances coverage and speci-
ficity.
\item \textbf{Semantic Deduplication:}
We compute dense
embeddings for each theme using BERT \cite{devlin2019bert}  and measure pairwise semantic similarity. For attributes with similarity exceeding
threshold $\tau$, we retain only the higher-frequency
variant, reducing semantic redundancy while preserving coverage.

\item \textbf{Human-in-the-Loop Validation:}
Human re-
view can play a complementary role in refining
the theme set by injecting domain and business
context that may be missed by fully automated
methods. In our setup, evaluators perform two
targeted actions. First, they consolidate semanti-
cally distinct but conceptually equivalent themes
(e.g., merging tour guide, host, and instructor).
Second, to address overly broad LLM-generated
themes such as Logistics, we leverage the fre-
quency distribution of extracted aspects within
each theme to identify cases where a generic
label is dominated by a small number of high-
volume aspect clusters. In such cases, evalua-
tors decompose the broad theme into more spe-
cific ones (e.g., promoting tour pacing and tour
itinerary to standalone themes), ensuring the fi-
nal theme set remains semantically distinct and
sufficiently granular to support customer evalua-
tion.

Importantly, this human-in-the-loop step is optional and selectively applied. To preserve scalability, only new themes that exceed a frequency
threshold and are not semantically similar to existing themes are flagged for review. In practice, the automated pipeline already produces
stable and usable themes, with human validation
primarily serving as an additional safeguard for
high-impact or domain-sensitive settings (e.g.,
regulated domains such as healthcare, or tour
experience). This design allows the system to
operate efficiently at scale while remaining flexi-
ble enough to incorporate expert oversight when
business or domain requirements warrant it.
\end{itemize}

\subsection{Theme-Constrained Opinion Extraction}
\label{subsec:theme-const-op-ext}

In this phase, we extract structured ABSA outputs (Figure~\ref{fig:absa_output}) by constraining the LLM to the consolidated theme set from Section~\ref{subsubsec:theme_refinement}. However, large theme sets and long prompts can degrade LLM performance \cite{hong2025context} and increase sensitivity to ordering of items in the prompt~\cite{guan2025order, choudhari2025prompt}. To address this, we adopt a two-stage extract-and-validate strategy: 
\paragraph{Recall Maximization (Shuffling)} To account
for LLM sensitivity to instruction ordering, we per-
form k = 3 few-shot prompt runs with randomly
shuffled theme definitions. The outputs from these
runs are collated, ensuring that themes missed in
one permutation due to position bias are captured
in another, thereby maximizing recall.
\paragraph{Precision Refinement (Binary Validation)} To
filter false positives from the aggregated output, we
introduce a binary validation step. Each extracted
tuple (theme, aspect, opinion, sentiment) is individ-
ually verified against the original review text and
the specific theme definition. Leveraging the strong
binary decision-making capabilities of LLMs \cite{liu2024aligning, arabzadeh2025benchmarking}, this
step effectively eliminates irrelevant extractions,
ensuring high precision in the final structured out-
put.

\subsection{Opinion-Aware Review Summarization}
Building on the structured ABSA outputs from Section~\ref{subsec:theme-const-op-ext}, we generate summaries using only theme-relevant opinions instead of full reviews, significantly reducing context length and improving the relevance of information presented to the LLM. Figure~\ref{fig:absa_output} illustrates this module using an example from the travel experience domain.

\begin{figure*}[ht]
    \centering
    \includegraphics[width=0.92\linewidth]{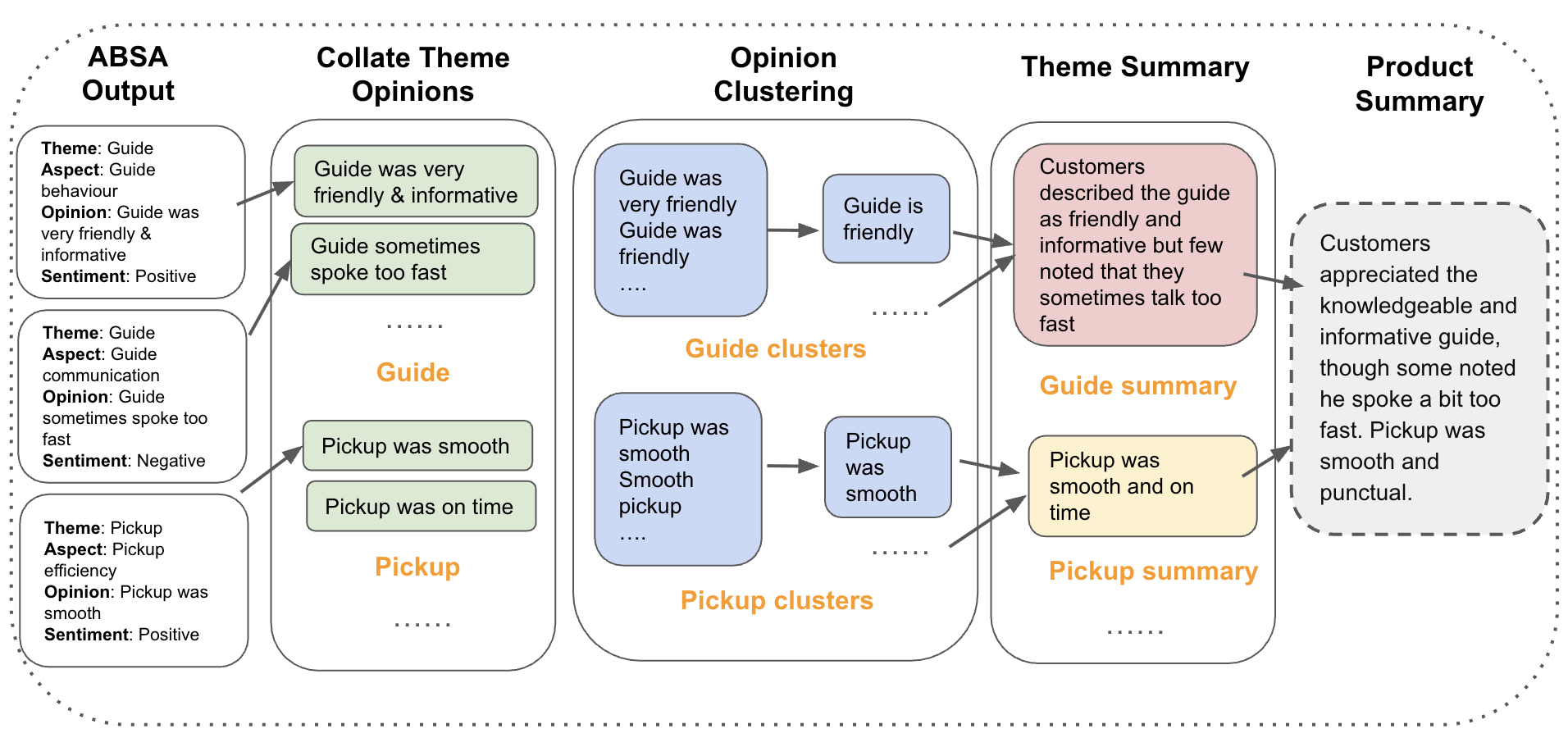}
    \caption{Overview of the proposed opinion‑summarisation pipeline using example from travel experience domain. Structured ABSA tuples (theme, aspect, opinion, sentiment) are collated by theme (e.g., “Guide”, “Pickup”), clustered to merge semantically similar opinions, and then summarised into concise theme‑level summaries, which are finally composed into a coherent product‑level summary.}
    \label{fig:input_output_train_pair}
\end{figure*}

\subsubsection{Opinion clustering}
Opinion extraction often produces substantial redundancy within a given theme~\cite{li2025decomposed}, which increases inference costs and can cause LLMs to hallucinate or omit salient minority viewpoints. To address this, we cluster opinions at the product-theme-sentiment level. For each cluster, we retain only three diverse, representative opinions to include in the summarization prompt, ensuring a compact and balanced context. We empirically demonstrate the impact of opinion clustering in Section \ref{sec: imp_op_clust}.

\label{subsubsec:opinion_clustering}

\subsubsection{Theme \& Product Summary} Once opinions are clustered, we aggregate the representative opinions to generate focused, \textbf{theme-level summaries}. This granular step ensures specific aspect nuances (e.g., Tour Guide vs. Pickup) are captured in detail without being overshadowed by dominant themes. In the final stage, the model synthesizes the theme summaries to generate a cohesive \textbf{product-level summary}. By operating on structured theme summaries rather than raw reviews, the final generation is both more relevant and easier to validate.
\section{Experimental Setup}
This section outlines the experimental setup, including the datasets, language models, and evaluation metrics, followed by results. Across all datasets, we perform opinion clustering at the product–theme–sentiment level using HDBSCAN \cite{rahman2016hdbscan}, with fixed hyperparameters $\texttt{cluster\_selection\_epsilon}=0.05$ and $\texttt{min\_samples}=5$ selected based on preliminary experiments. For both theme-level and product-level summarization, we employ a few-shot LLM prompting strategy. The prompts used in our experiments are provided in Appendix~\ref{section:prompts_appendix} while the intermediate results and insights from Theme Discovery \& Standardization on SPACE is summarized in the Appendix \ref{sec: space_theme_ref_and_consol}. For experiments on the TRECS dataset, we used GPT-4.1 for theme extraction and GPT-4o/Llama-3.1-70B for summarization.

\subsection{Datasets}

\subsubsection{Public Datasets}

We conducted experiments on 2
domains, specifically  business reviews for hotels (SPACE) and scientific reviews for research articles (PeerSum). 
For hotels, we conduct experiments on the SPACE dataset \cite{angelidis2021extractive}, as it is the only large-scale publicly available dataset that provides both theme-level and product-level summaries in a unified setting. PeerSum \cite{li2023summarizing} contains reviews for scientific articles and corresponding meta-reviews from OpenReview. 
We used the same test datasets as \cite{li2025decomposed} and report the metrics of the various methods presented in that paper for comparison with MOSAIC.

\subsubsection{TRECS}

To support further research on reviews and short-document summarization, we release TRECS, an open-source dataset from the travel experience domain. The dataset comprises 140,631 reviews spanning 344 tour products, with an average of approximately 400 reviews per product and a mean review length of 280 characters.
For each product, we identify 8 frequently discussed themes extracted from the reviews, resulting in 36 unique themes across the dataset, which is at least 5x as compared to the datasets like SPACE and PeerSum. 
Theme-level summaries are generated for each identified theme, and these are reused to construct product-level summaries. This hierarchical approach ensures alignment and consistency between theme and product-level summaries, directly addressing the gap identified in Section~\ref{sec:space_data_deepdive}. All summaries have been verified and corrected by a team of human annotators. Additional details on dataset construction are provided in Appendix~\ref{section:trecs_appendix}, and statistics are shown in Appendix Table~\ref{tab:dataset_statistics}.
During experimentation, GPT-4.1 was used for theme extraction, and GPT-4o and Llama-3.1-70B for summarization.

\subsection{Evaluation Metrics}
\label{subsection: eval_metrics}
We evaluate the quality of product summaries along the following two dimensions: 

\noindent\textbf{Aspect coverage} evaluates how well a generated summary captures the themes discussed in the reviews, computed as the F1 score between themes identified in the summary and those in the source reviews, using the same theme-identification prompt as \cite{li2025decomposed}. 

\noindent\textbf{Faithfulness} measures the extent to which summary opinions are supported by the reviews, evaluated using G-Eval \cite{liu2023g} and AlignScore \cite{zha2023alignscore}, a fine-tuned alignment-based metric (with large backbone, \texttt{nli\_sp} mode). We always report average scores over three runs. 
\section{Results}
In this section we discuss the results for online and offline evaluation of our framework MOSAIC.
\begin{table*}[ht]
\centering
\footnotesize
\begin{tabular}{lcccc}
\hline
\textbf{Methods} & \textbf{Coverage$\uparrow$} & \textbf{G-Eval$\uparrow$} & \textbf{AlignScore-R/M$\uparrow$} \\
\hline
Sentiment CoT-GPT-4o \cite{li2024sentiment}  & 0.96 & 0.75 & 0.72 / 0.08 \\
FT-Llama 8B \cite{touvron2023llama} & 0.87 & 0.60 & 0.33 / 0.06\\
Aspect-aware decomposition-GPT-4o \cite{li2025decomposed} & 0.95 & 0.76 & 0.68 / 0.06\\
MOSAIC-GPT-4o (ours) & \underline{\textbf{0.99}} & \underline{\textbf{0.84}} & \underline{\textbf{0.81}} / \underline{0.16}\\   
\hline
Automatic decomposition-Llama 70B \cite{khot2022decomposed} & 0.59 & 0.31 & 0.51 / 0.03\\
Chunk-wise decomposition-Llama 70B \cite{khot2022decomposed} & 0.84 & 0.72 & 0.65 / 0.06 \\
Naive aspect-aware prompting-Llama 70B \cite{radford2019language} & 0.72 & 0.62 & 0.70 / 0.07\\
Aspect-aware decomposition-Llama 70B \cite{li2025decomposed} & 0.97 & 0.76 & 0.76 / 0.09\\
MOSAIC-Llama 70B (ours) & \underline{\textbf{0.99}} & \underline{0.82} & \underline{\textbf{0.81 / 0.19}}\\
\hline
\end{tabular}
\caption{Results on PeerSum - The first section of the table presents results for
GPT-4o and state-of-the-art models. The second section has results for Llama 70B.
Underlined scores denote best in section per metric while bold scores denote best overall. AlignScore-R calculates
AlignScore against source reviews, while AlignScore-M is computed against reference meta-reviews.}
\label{tab:peersum_comparison_table}
\end{table*}

\subsection{Online Evaluation}

We validated the practical utility of our intermediate outputs through online A/B tests on a major travel experience platform:

\begin{itemize}[leftmargin=*]
    \item \textbf{Review Sorting:}
    Review sort driven by theme and sentiment level outputs from Section~\ref{subsubsec:unconstrained_theme_gen} resulted in a statistically significant 1\% uplift in conversion rate ($p < 0.1$), suggesting that sentiment-aware ordering of reviews can meaningfully improve user engagement and purchasing behavior.
    
    \item \textbf{Interactive Review Themes:} Surfacing review themes as clickable filters on product pages (see Figure~\ref{fig:industrial_application}, ``Why travelers loved this'') led to a statistically significant 1.5\% increase in revenue per visitor ($p < 0.1$), suggesting that improved review transparency can encourage higher-value purchases.

    \item \textbf{Traveler Tips:} We are currently running online A/B tests evaluating structured, review-derived actionable advice for future travelers (see Figure~\ref{fig:industrial_application}, ``Tips from satisfied travellers''). Early results show promising directional improvements in user engagement and revenue related metrics.
\end{itemize}

\subsection{Offline Evaluation}
We generate summaries for all datasets using MOSAIC and compare them against strong baselines and state-of-the-art methods, relying on published results from \cite{li2025decomposed} obtained under identical evaluation settings.

\subsubsection{Public Datasets}
Tables \ref{tab:peersum_comparison_table} and Appendix Table~\ref{tab:space_comparison_table}
 summarize the results on both the datasets PeerSum and SPACE respectively on different evaluation metrics.

\noindent\textbf{PeerSum (Table \ref{tab:peersum_comparison_table})}. Our method outperforms the strongest prior baseline across all reported metrics. 
With GPT-4o, we observe clear relative improvements in Coverage and G-Eval, alongside substantially larger gains in AlignScore-R ($\sim$19\%) and AlignScore-M (over $2\times$), indicating markedly improved factual consistency. 
For Llama-70B, gains are smaller but consistent, with modest improvements in Coverage and G-Eval, and relative increases of $\sim$6–7\% in AlignScore-R and $\ge$100\% in AlignScore-M.

\noindent\textbf{SPACE (Table \ref{tab:space_comparison_table})}. Under GPT-4o, both methods achieve identical Coverage and G-Eval, while \cite{li2025decomposed} attains slightly higher AlignScore-R/M. 
This trend reverses for Llama-70B, where our method improves Coverage and yields sizeable relative gains in AlignScore-R ($\sim$7–8\%) and AlignScore-M ($\sim$45\%), demonstrating stronger factual consistency under a weaker base model.
\begin{table*}[ht]
\centering
\footnotesize
\begin{tabular}{lcccc}
\hline
\textbf{Methods} & \textbf{Coverage$\uparrow$} & \textbf{G-Eval$\uparrow$} & \textbf{AlignScore-R/M$\uparrow$}\\
\hline
HIRO-abs \cite{hosking2024hierarchical} & 0.87 & 0.62 & \underline{0.83}/\underline{\textbf{0.24}}\\
TCG \cite{bhaskar2022prompted} & 0.98 & 0.66 & 0.66/0.11\\
Aspect-aware decomposition-GPT-4o \cite{li2025decomposed} & \underline{\textbf{1.00}} & \underline{\textbf{0.90}} & 0.81/0.10\\
MOSAIC GPT-4o (ours) & \underline{\textbf{1.00}} & \underline{\textbf{0.90}} & 0.77/0.07\\
\hline
Automatic decomposition-Llama 70B \cite{khot2022decomposed} & 0.63 & 0.38 & 0.70/\underline{0.22}\\
Chunk-wise decomposition-Llama 70B \cite{khot2022decomposed} & 0.93 & 0.84 & 0.65/0.01\\
Naive aspect-aware prompting-Llama 70B \cite{radford2019language} & 0.37 & 0.34 & 0.44/\underline{0.22}\\
Aspect-aware decomposition-Llama 70B \cite{li2025decomposed} & \underline{0.99} & \underline{0.88} & 0.79/0.11\\
MOSAIC-Llama 70B (ours) & \underline{0.99} & \underline{0.88} & \textbf{0.86}/0.16\\
\hline
\end{tabular}

\vspace{2mm}
\caption{Results on SPACE - The first section of the table presents results for GPT-4o and state-of-the-art models. The second section has results for Llama 70B. Underlined scores denote best in section per metric, while bold scores denote best overall. AlignScore-R calculates AlignScore against source reviews, while AlignScore-M is computed against reference meta-reviews.}
\label{tab:space_comparison_table}
\end{table*}

Across both PeerSum and SPACE, MOSAIC consistently matches or outperforms the strongest prior methods in coverage and alignment metrics, demonstrating robust gains in summary faithfulness across domains and models.


\begin{table}[t]
\centering
\footnotesize
\begin{tabular}{lccc}
\hline
 & 
 \textbf{G-Eval$\uparrow$} & \textbf{AlignScore-R/M$\uparrow$} \\
\hline
Without OC (GPT-4o) &  \underline{\textbf{0.859}} & 0.904/0.48\\
With OC (GPT-4o) & 0.858 & \underline{0.909}/\underline{0.518}\\
\hline
Without OC (Llama 70B) & 0.631 & \textbf{0.935}/0.546\\
With OC (Llama 70B) & \underline{0.684} & 0.933/\underline{\textbf{0.575}}\\
\hline
\end{tabular}
\caption{Results on TRECS dataset. First section of the table presents results with and without opinion clustering (OC) and using GPT-4o for summarization, and second section has results for Llama3.1-70B.
}

\label{tab:viator_comparison_table}
\end{table}

\subsubsection{TRECS}
Table \ref{tab:viator_comparison_table} highlights the results on TRECS dataset along with highlighting the effect of opinion clustering. 
Across both GPT-4o and Llama3.1-70B, incorporating opinion clustering yields consistent gains in alignment-based metrics while preserving high coverage and summarization quality. For GPT-4o, opinion clustering improves AlignScore-M by 7.9\% and AlignScore-R by 0.6\%, and we observe no meaningful change in G-Eval. 
For Llama-3.1-70B, opinion clustering leads to an 8.4\% increase in G-Eval and a 5.3\% improvement in AlignScore-M.

Overall, these results indicate that opinion clustering improves factual consistency and meta-review alignment without degrading summary completeness; however, its full impact may be understated due to the limited redundancy in the dataset. 
This motivates our targeted stress tests in Section~\ref{sec: imp_op_clust}, which reveal that LLMs are sensitive to opinion ordering in the prompt (shuffled vs. ordered). Crucially, opinion clustering neutralizes this volatility, ensuring robust performance regardless of input structure.
\section{Importance of Opinion Clustering}
\label{sec: imp_op_clust}
\begin{figure}[!ht]
    \centering
    \begin{subfigure}
        \centering
        \includegraphics[width=0.95\linewidth]{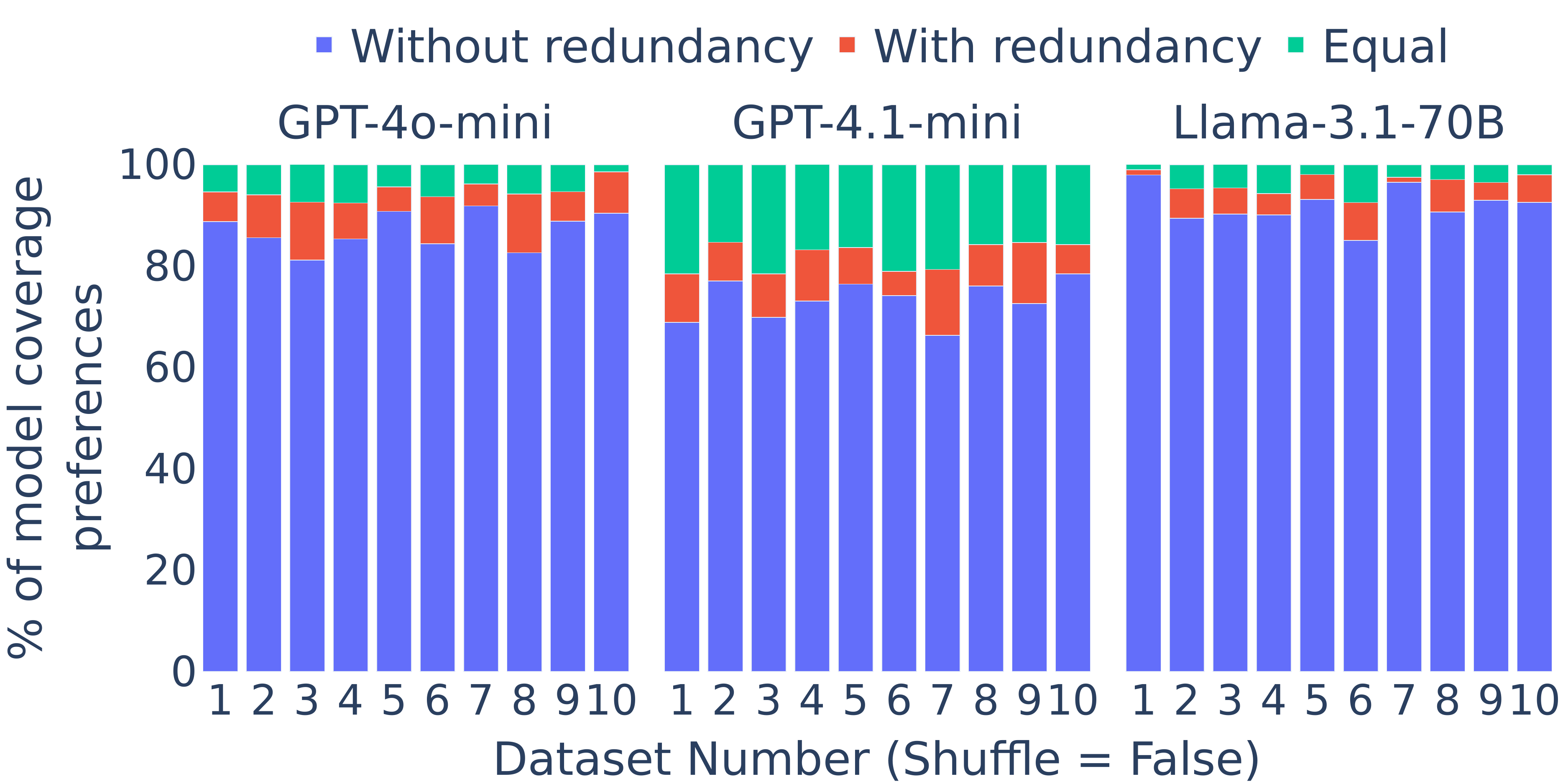}
        \label{fig:coverage_shuffle_false}
    \end{subfigure}
    \hfill
    \begin{subfigure}
        \centering
    \includegraphics[width=0.95\linewidth]{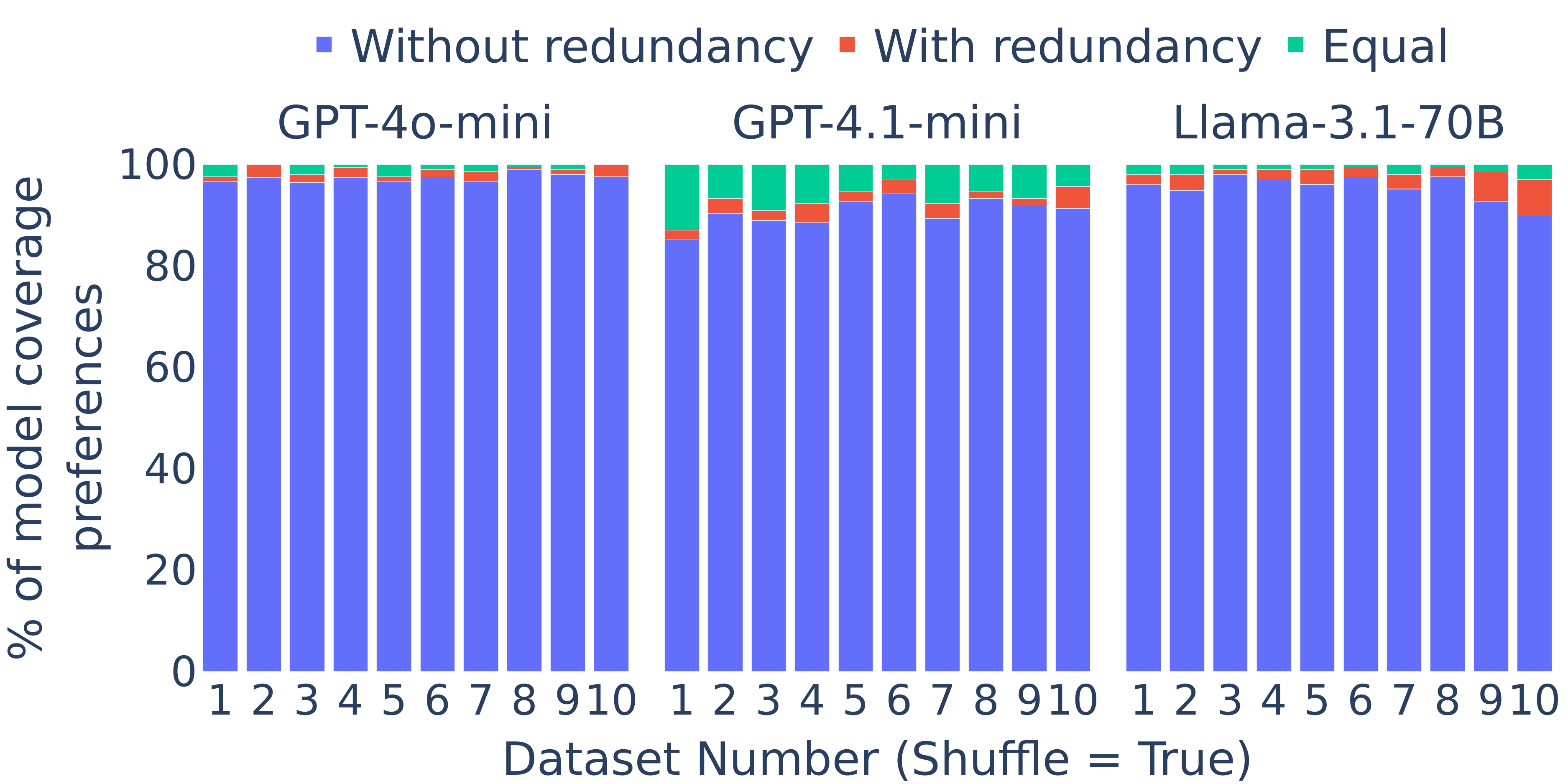}
        \label{fig:coverage_shuffle_true}
    \end{subfigure}
    \caption{Preference of three LLMs (GPT-4o-mini, GPT-4.1-mini, and Llama-3.1-70B) when selecting between summaries generated from redundant opinions and those generated from deduplicated opinions, based on aspect coverage across 10 datasets. For each dataset (X-axis), bars represent the percentage of model responses favoring the non-redundant summary (blue), the redundant summary (red), or indicating equal coverage (green).} 
    \label{fig:coverage_comparison}
\end{figure}

\begin{figure}[!ht]
    \centering
    \begin{subfigure}
        \centering
    \includegraphics[width=0.95\linewidth]{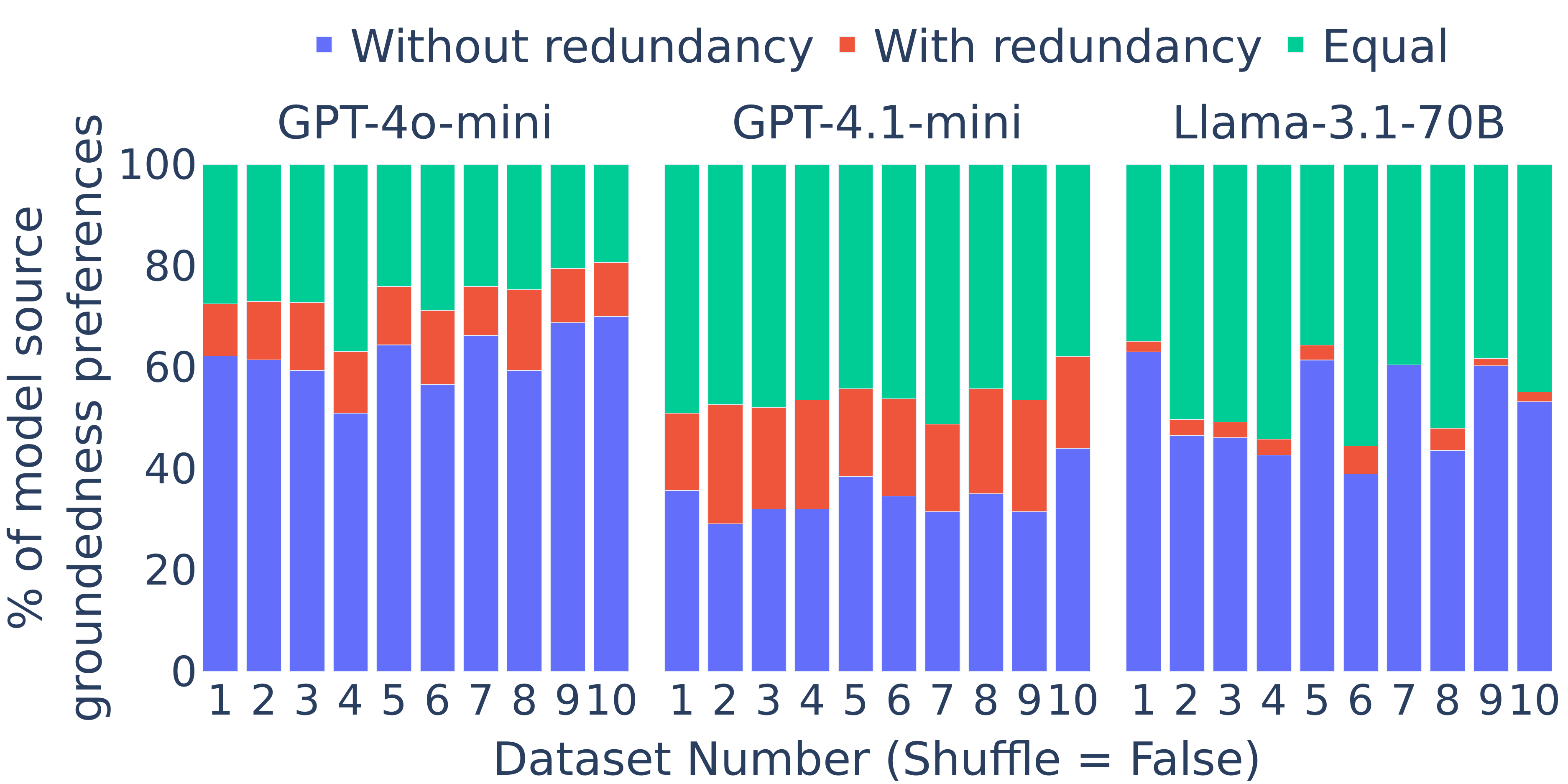}
        \label{fig:faithfulness_shuffle_false}
    \end{subfigure}
    \begin{subfigure}
        \centering
        \includegraphics[width=0.95\linewidth]{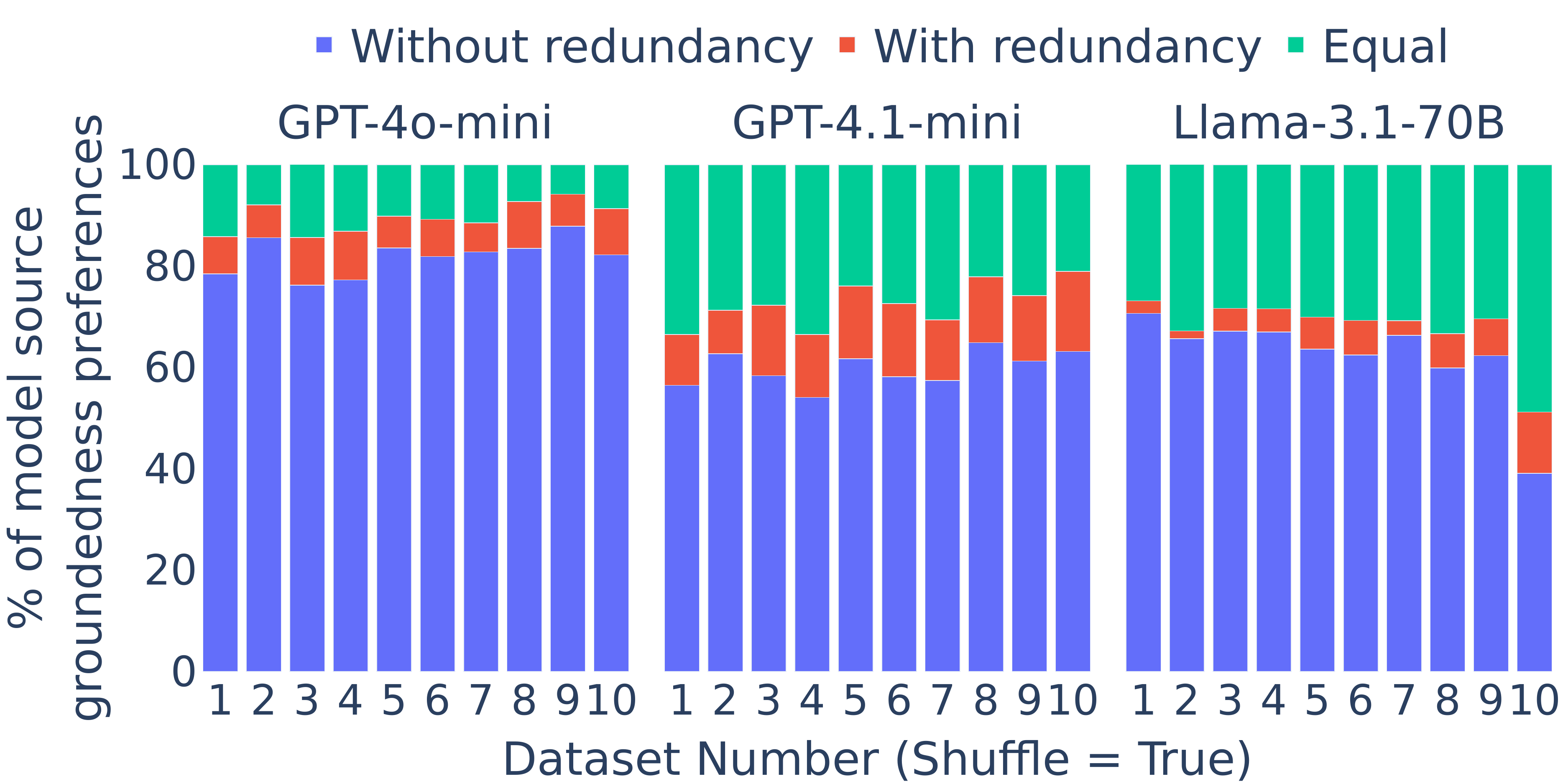}
        \label{fig:faithfulness_shuffle_true}
    \end{subfigure}

    \caption{Preference of three LLMs (GPT-4o-mini, GPT-4.1-mini, and Llama-3.1-70B) when selecting between summaries generated from redundant opinions and those generated from deduplicated opinions, based on aspect faithfulness across 10 datasets. For each dataset (X-axis), bars represent the percentage of model responses favoring the non-redundant summary (blue), the redundant summary (red), or indicating equal coverage (green).}
    \label{fig:faithfulness_comparison_appendix}
\end{figure}

While the TRECS dataset represents real-world distribution, it does not fully capture the extreme opinion redundancy found in high-volume product pages (e.g., thousands of reviews repeating the same point). 
To strictly isolate and stress-test  the benefits of our clustering module under these high-noise conditions, we constructed a controlled synthetic evaluation benchmark that systematically introduces varying degrees and patterns of redundancy.
This benchmark comprises 10 dataset variants, each designed to expose different redundancy intensities and positional effects. Details of the dataset construction are provided in Appendix~\ref{sec: opinion_redundancy_data_appendix}.
This is specifically designed to surface large-scale opinion redundancy -- a challenge not captured by existing public datasets. 

For each synthetic dataset variant, we generate theme-level summaries under two conditions: (i) with redundant opinions retained and (ii) after redundancy removal via opinion clustering. 
To evaluate the impact of redundancy on summarization quality, we employ an LLM-as-a-judge to perform pairwise comparisons between the two summaries, producing a preference or tie decision along the following dimensions:
\begin{itemize}[leftmargin=*]
\item \textbf{Content Coverage}: whether the summary captures all key points expressed in the opinion text (e.g., guide friendliness, informativeness, crowd management).
\item \textbf{Source Groundedness}: whether the summary avoids introducing information not supported by the source opinions.
\end{itemize}

To further analyze the role of input ordering, we evaluate each dataset variant under two prompt configurations: one where redundant opinions are arranged such that similar opinions are grouped together, and another where opinions are randomly shuffled.
Figure~\ref{fig:coverage_comparison}  reports results for both settings - unshuffled(top) and shuffled(bottom), 
highlighting how redundancy severity and ordering jointly affect summarization performance.
Figure \ref{fig:faithfulness_comparison_appendix} shows the results for both settings unshuffled(top) and shuffled(bottom), highlighting the impact of opinion ordering on source groundedness.
Following three key insights are reflected from the analysis:
\begin{itemize}[leftmargin=*]
    \item Removing redundant opinions from the input substantially boosts both coverage and faithfulness in generated summaries regardless of the summarization model used.
    \item When opinion order is shuffled in the prompt, the negative impact of redundant opinions on both coverage and faithfulness becomes more pronounced.
    \item The GPT-4.1-mini model demonstrates higher consistency in producing summaries with equal coverage across varying redundancy and shuffling conditions, outperforming both Llama-3.1-70B-Instruct and GPT-4o-mini on this metric.
\end{itemize}
These results underscore the importance of opinion deduplication and opinion structuring in the prompt as critical preprocessing steps for robust and faithful summarization in real-world settings, where redundancy and input order are uncontrolled.

\section{SPACE Deep Dive}
\label{sec:space_data_deepdive}

As noted in \cite{li2025decomposed}, a high-quality summary should accurately reflect the balance of opinions in the source reviews and address the entity’s specific themes (e.g., Cleanliness, Service, Location). Building on this, we additionally examine theme coverage in the SPACE dataset by comparing our generated theme summaries with the corresponding theme summaries provided in SPACE using two key evaluation metrics:
\begin{itemize}[leftmargin=*]
\item \textbf{Sentiment Score}:
We apply a zero-shot approach to assign a sentiment score in the range [0, 100] to each theme summary, where higher scores indicate more positive sentiment.
\item \textbf{Theme Coverage}: We measure theme coverage by counting the number of distinct themes mentioned in the product summary. 

\end{itemize}


\begin{figure}[!ht]
    \centering
    \includegraphics[width=0.9\linewidth]{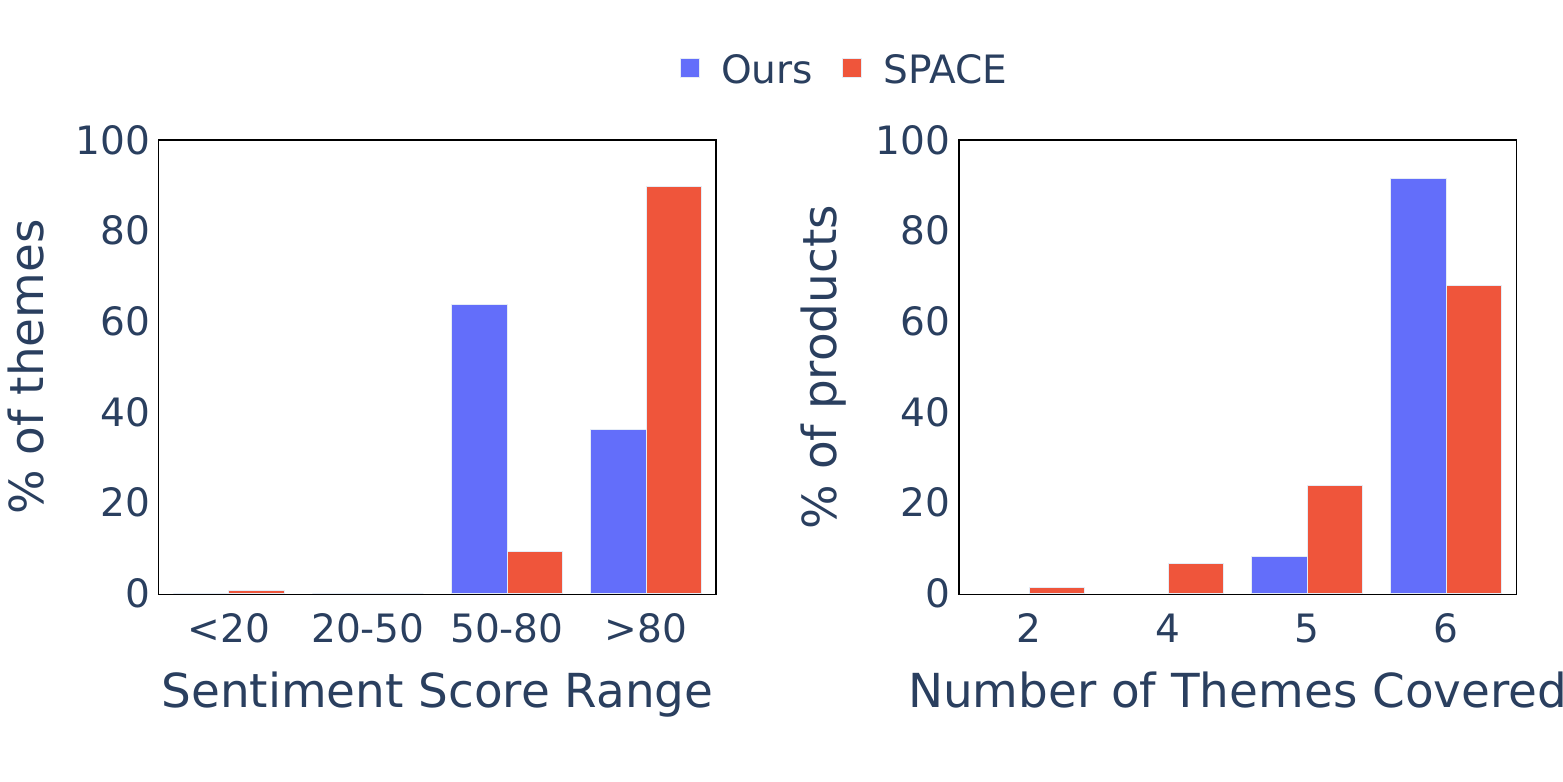}
    \caption{Distribution of sentiment scores for theme-level summaries and of the number of themes identified in product summaries for SPACE and MOSAIC, both evaluated using Llama-3.1‑70B}
    \label{fig:space_deepdive_fig}
\end{figure}
\noindent\textbf{Key insights}
\begin{itemize}[leftmargin=*]
\item As shown in Figure~\ref{fig:space_deepdive_fig}, SPACE summaries are heavily skewed toward highly positive sentiment (\textgreater 80), whereas MOSAIC produces a larger proportion of neutral summaries (50–80), indicating reduced positivity bias. To validate this, we conducted human evaluation on a random sample of 25 summaries from the 50–80 and \textgreater 80 bins where the two models disagreed. MOSAIC correctly identified overall theme sentiment in all 25 cases (validated manually).
\item As shown in Figure~\ref{fig:space_deepdive_fig}, MOSAIC produces product summaries that cover all six themes in 92\% of cases, compared to 69\% for SPACE, indicating substantially higher completeness in theme coverage.
\end{itemize}

This raises concerns about the quality of the SPACE dataset, which is widely used in the research community, and highlights an important implication: human annotators should be provided with clear guidance on which themes to include, ensuring proper alignment between theme and product summaries. Additional analysis and insights
are provided in Appendix \ref{appsec:space_data_deepdive}.
\section{Future Work \& Conclusion}
While MOSAIC improves faithfulness, transparency, and user engagement, it has several limitations. The quality of intermediate outputs (e.g., aspect and opinion extraction) depends on the capabilities of the underlying LLM, and less capable models may struggle with nuanced or implicit expressions. Its modular design also requires domain-specific prompt engineering and clustering hyperparameters, which may not generalize uniformly across datasets. More fundamentally, generating a balanced summary under length constraints is challenging, as it requires weighing common opinions against less frequent but potentially important viewpoints, which may vary across domains and user needs. This highlights the need for principled metrics capturing coverage, balance, and fairness in opinion summarization.

These limitations motivate several directions for future work. The pipeline could be fully automated using an agentic framework to adaptively select prompts and validate intermediate outputs. Extensions could support personalized summarization by prioritizing themes based on user preferences, mitigate repetitive tones in AI-generated summaries, enable multilingual generalization \cite{wu2025m}, or fine-tune task-specific models for aspect and opinion extraction.

In this work, we present MOSAIC, a scalable, modular framework that bridges offline benchmarks and real-world applications. By decomposing summarization into interpretable components, MOSAIC addresses noise and redundancy in large-scale reviews, and its intermediate outputs have practical value as demonstrated in online A/B experiments on a travel marketplace, showing measurable improvements in user engagement and revenue. Offline evaluations further confirm that MOSAIC consistently outperforms state-of-the-art methods across multiple domains, with opinion clustering enhancing summary faithfulness. To support reproducibility and future research, we publicly release the TRECS dataset and the full codebase, enabling further work on robust, transparent, and experience-centric review summarization.

\section{Ethics Statement}
Our work seeks to enhance customer experience by making product reviews more accessible through structured and transparent summarization. 
Beyond generating summaries, the system exposes interpretable intermediate outputs, enabling customers to trace conclusions back to supporting evidence and form their own informed judgments. 
These components are designed to augment rather than to replace human decision making by reducing the cognitive burden of processing large volumes of reviews. 
We release the code, model, and a dataset (TRECS) to encourage further research aimed at identifying and mitigating unwanted biases, thereby contributing to broader societal benefit.

\bibliography{references}

\appendix
\onecolumn
\section{Dataset}
\subsection{Statisics}
\begin{table}[ht]
\centering
\fontsize{9}{11}\selectfont
\setlength{\tabcolsep}{0.8pt}
\begin{tabular}{p{1.6cm} p{2.6cm} p{1.4cm} p{1cm} p{1cm}}
\hline
\textbf{Dataset} &
\textbf{\#Train / Dev / Test} &
\textbf{\#Reviews} &
\textbf{MetaL} &
\textbf{\#Aspects} \\
\hline
PeerSum & 22,420 / 50 / 100 & 14.9 & 156.1 & 5 \\
SPACE   & 0 / 25 / 25  & 100   & 75.7  & 6 \\
TRECS & 14,0631 & 409 & 94.8  & 36 \\
\hline
\end{tabular}
\caption{Statistics of our experimental datasets. \#Train/Dev/Test denote the number of training, development, and test instances; \#Reviews is the average number of reviews per entity; MetaL to the average meta-review length; \#Aspects denotes the number of aspects covered.}
\label{tab:dataset_statistics}
\end{table}

\subsection{TRECS}
\label{section:trecs_appendix}

We constructed the TRECS dataset by applying MOSAIC to generate both theme-level summaries and product-level summaries from raw reviews, followed by human validation. Theme discovery and standardization were performed using GPT-4o-mini, while GPT-4.1-mini was used for both theme-constrained opinion extraction and opinion-aware review summarization. All generated theme and product summaries were then evaluated by a team of human annotators to ensure quality and consistency.
Table~\ref{tab:dataset_stats_TRECS} shows the distribution of top 10 themes in the data set.

\begin{table}[h]
\centering
\label{tab:theme_counts}
\begin{tabular}{l r}
\hline
\textbf{Theme} \& \textbf{Count} \\
\hline
Tour Guide or Instructors        & 320 \\
Tour Informativeness             & 287 \\
Tour Planning \& Coordination    & 252 \\
Point of Interest                & 237 \\
Natural Landscape Viewing        & 174 \\
Food \& Beverage                 & 172 \\
Driver                           & 169 \\
Tour Activities \& Participation & 160 \\
Value for Money                  & 110 \\
Tour Pacing                      & 110 \\
\hline
\end{tabular}
\caption{Distribution of the top 10 most frequent themes in the TRECS dataset. This illustrates the thematic diversity of the travel experience domain, with 'Tour Guide' and 'Tour Informativeness' being the most discussed categories.}
\label{tab:dataset_stats_TRECS}
\end{table}

\subsection{Opinion Redundancy Dataset}
\label{sec: opinion_redundancy_data_appendix}

To systematically evaluate the impact of opinion redundancy on theme-level opinion summarization, we constructed a controlled dataset that introduces varying levels and patterns of redundancy into opinion collections. Our dataset construction pipeline consists of four main stages:
\subsection{Cluster Quality Filtering}
\label{subsubsec: cluster_quality_filtering}
Following MOSAIC, we process reviews from a major travel experience platform and use the output of Section~\ref{subsubsec:opinion_clustering} (opinion clustering) as input to this stage. To ensure that only semantically coherent opinion groups are retained, we apply automated cluster quality filtering based on two complementary cohesion metrics:
\begin{itemize}[leftmargin=*]
    \item \textbf{Average distance from centroid} $\leq 0.2$, ensuring that opinions within a cluster are tightly concentrated around the centroid.
    \item \textbf{Average pairwise similarity} $\geq 0.7$, enforcing high inter-opinion coherence within each cluster.
\end{itemize}
Clusters failing either criterion are discarded. Filtering is performed at the \textit{Product-Theme-Sentiment} granularity, ensuring that subsequent sampling draws from high-quality, semantically consistent opinion groups.

\subsection{Diverse Opinion Sampling}
\label{subsubsec: diverse_opinion_sampling}
After identifying high-quality clusters, we consider the unclustered opinions within each \textit{Product–Theme–Sentiment} group, i.e., those not assigned to any HDBSCAN cluster. From this pool of unclustered opinions, we employ Maximal Marginal Relevance~\cite{carbonell1998use} sampling to select a diverse set of opinions. MMR balances two objectives:
\begin{itemize}[leftmargin=*]
    \item \textbf{Relevance}: Selecting opinions that are central to the theme.
    \item \textbf{Diversity}: Avoiding near-duplicate or semantically overlapping opinions.
\end{itemize}

Formally, the MMR score for an opinion $i$ is computed as:
\[
\text{MMR}(i) = (1 - \lambda)\,\text{Relevance}(i) - \lambda \max_{j \in S} \text{Similarity}(i, j),
\]
where $\lambda = 0.8$ controls the diversity-relevance trade-off, $S$ is the set of already selected opinions, relevance is defined as the mean cosine similarity of $i$ to all other opinions in the group, and similarity is the pairwise cosine similarity between opinion embeddings. Opinion embeddings are generated using the \texttt{all-MiniLM-L6-v2} sentence transformer. For each \textit{Product--Theme--Sentiment} group, we sample up to 10 opinions via MMR.

\subsection{Strategic Opinion Selection}
\label{subsubsec:strategic_opinion_selection}
To construct a fixed-size base opinion set for each \textit{Product--Theme--Sentiment} group, we combine opinions from two complementary sources:
\begin{itemize}[leftmargin=*]
    \item \textbf{Cluster representatives}: One opinion randomly selected from each high-quality cluster retained after cluster quality filtering (Section~\ref{subsubsec: cluster_quality_filtering}).
    \item \textbf{Diverse unclustered opinions}: Opinions sampled via Maximal Marginal Relevance (MMR) from the remaining unclustered opinion pool (Section~\ref{subsubsec: diverse_opinion_sampling}).
\end{itemize}

The selection process prioritizes cluster representatives to ensure that dominant semantic clusters are explicitly represented. If the number of cluster representatives is fewer than the target size, we fill the remaining slots using MMR-selected unclustered opinions, up to a maximum of 10 opinions per group. This strategy ensures that each \textit{Product--Theme--Sentiment} group contains exactly 10 base opinions, enabling controlled and systematic redundancy injection in subsequent stages.

\subsection{Redundancy Pattern Generation}
\label{subsubsec:redundancy_pattern_generation}

Given the fixed-size base opinion set of 10 opinions constructed for each \textit{Product--Theme--Sentiment} group (Section~\ref{subsubsec:strategic_opinion_selection}), we systematically introduce opinion redundancy by duplicating selected opinions within this set.

Specifically, we define 10 redundancy patterns, each represented by a duplication count vector
$\mathbf{c} = [c_1, c_2, \ldots, c_{10}]$, where $c_i$ denotes the number of times the $i$-th base opinion is repeated in the final input. The ordering of opinions follows the order established during strategic opinion selection, ensuring consistency across patterns.

Each redundancy pattern is designed to vary along 2 key dimensions: (i) redundancy intensity (ranging from moderate to extreme duplication), (ii) redundancy position (start, middle, end, or scattered within the opinion sequence) (see Table~\ref{tab:duplication_patterns}).

\begin{table}[t]
\centering
\small
\begin{tabular}{c c p{7.5cm}}
\hline
\textbf{Dataset Number} & \textbf{Duplication Vector} & \textbf{Description} \\
\hline
1  & {[}3000, 5, 5, 5, 5, 5, 5, 5, 5, 3000{]} & Extreme redundancy at boundaries \\
2  & {[}5, 5, 5, 3000, 3000, 5, 5, 5, 5, 5{]} & Extreme redundancy in the middle \\
3  & {[}3000, 3000, 5, 5, 5, 5, 5, 5, 5, 5{]} & Extreme redundancy at the start \\
4  & {[}3000, 5, 5, 3000, 5, 5, 5, 5, 5, 5{]} & Extreme redundancy scattered across positions \\
5  & {[}2000, 2000, 5, 5, 5, 5, 5, 5, 5, 2000{]} & High redundancy at boundaries \\
6  & {[}2000, 2000, 2000, 5, 5, 5, 5, 5, 5, 5{]} & High redundancy at the start \\
7  & {[}5, 5, 5, 5, 5, 5, 5, 2000, 2000, 2000{]} & High redundancy at the end \\
8  & {[}2000, 2000, 5, 5, 2000, 5, 5, 5, 5, 5{]} & High redundancy scattered across positions \\
9  & {[}5, 5, 5, 5, 2000, 5, 5, 5, 2000, 2000{]} & High redundancy at end positions \\
10 & {[}1000, 1000, 1000, 1000, 5, 5, 5, 5, 5, 2000{]} & Mixed redundancy levels \\
\hline
\end{tabular}
\caption{Synthetic redundancy patterns designed to stress-test model robustness with respect to opinion redundancy. The vectors specify extreme duplication counts at various positions (start, middle, end, or scattered) to isolate the effects of redundancy and input ordering on summarization quality.}
\label{tab:duplication_patterns}
\end{table}

For each \textit{Product--Theme--Sentiment} group, we generate 10 dataset variants-one per redundancy pattern resulting in controlled inputs that isolate the effects of redundancy intensity, position, and distribution on downstream summarization performance.




\section{SPACE Deep dive}
\label{appsec:space_data_deepdive}
We further identified the top six themes for each of the 50 products in the SPACE dataset based on opinion frequency and compared them with the six static themes used in SPACE. The results are summarized in Table~\ref{tab:additional_top6_themes}.

\begin{itemize}
    \item Using our approach, 43 out of 50 products exhibit at least one top-six theme that is not covered by SPACE’s fixed theme set, highlighting the limitations of applying the same static themes across all product summaries.
    \item Value for money appears among the top six themes for 25 out of 50 products.
    \item 24 product summaries explicitly reference the value for money/pricing aspect, which complicates evaluation, as this aspect is absent from SPACE’s predefined themes and can negatively affect metrics such as ROUGE and Alignscore-M. Some of the examples are provided in table~\ref{tab:value_for_money_summaries_space}
\end{itemize}

\begin{center} 
\small
\setlength{\tabcolsep}{6pt}
\renewcommand{\arraystretch}{1.3}
\begin{longtable}{|c|p{4.7cm}|p{4.7cm}|p{4.7cm}|}
\hline
\textbf{ID} & \textbf{v1} & \textbf{v2} & \textbf{v3} \\
\hline
\endfirsthead
\hline
\textbf{ID} & \textbf{v1} & \textbf{v2} & \textbf{v3} \\
\hline
\endhead
1 &
The staff are friendly and exceptional. Every room (lobby included) was very clean. They are spacious, very quiet, and come with a coffee maker. Though, the rooms are outdated in decor. The hotel itself is conveniently close to the airport and restaurants. There’s a chocolate-chip cookie at arrival, and for the prices, \hl{the experience is a good value}. &
Service was exceptional and the quality was great! The rooms are always clean, quiet and spacious with nicely appointed bathrooms. The location is across the street from the airport, was within walking distance to a Denny’s and other restaurants. The hotel interior itself is a bit outdated, but the room we stayed was modern. &
All the staff was exceptionally helpful, courteous, and friendly, keeping the rooms clean and well-prepared. The interior of the hotel needs updating, but the rooms themselves were very spacious, modern, and comfortable to stay in. The hotel itself is conveniently located near the airport, a steak restaurant, fast food, and has a free shuttle service for broader access to Seattle. \\
\hline
2 &
Hotel staff were all wonderfully accommodating and friendly. The entire hotel was very clean and the rooms even smell “fresh”. The rooms were large, with large beds, and modern, essential amenities that made it feel cozy. The swim-up pool bar has great, grilled food and reasonably priced drinks; and the continental breakfast was plentiful. (There are not enough reviews available to mention the location.) &
The staff was helpful and friendly, recommending many places to eat. The entire property is very clean, and the room was clean and comfy. The rooms are large, with comfortable beds, and plenty of amenities. The breakfast was plentiful and great, as was the food at the swim-up bar and grill. The location is excellent, close to the airport, but the pool was a bit small, and loud from the planes. It has free WiFi, and \hl{overall the hotel is an incredible value}. &
The staff was very professional and helpful to us. The hotel was very clean and well kept. The rooms were spacious and comfortable. There is a continental breakfast buffet included is great and there is a swim-up bar and grill which has very good food, too. The location is excellent but the Pool was a bit loud from the planes as it is near the airport. \\
\hline
3 &
Service was excellent; They were helpful, friendly, and gave great tips on where to go \& how to get there. Our room was very nice and large, with a lovely bathroom. It even had a small provision for skylight. Breakfast was ample and delicious if you like the typical European breakfast. The hotel was situated in a great location, in a calm area away from the busy parts of the city, but still comfortable walking distance to everything. \hl{It’s a very good value for price}. &
The staff were pleasant and helpful, especially when pointing out places of interest. Everything was very clean; the rooms even smelled good. The rooms as a whole were very large, with a lovely bathroom, and comfortable enough for a good night’s sleep. The European style breakfast menu was ample and delicious. The hotel was in a quiet location, but close to the Museum of Natural History and within walking distance to everything else. &
Staff were very friendly and helpful to show us the direction to places of interest and restaurants. Overall the room- Smelled good, clean. The bathroom is newly renovated and very clean and functional. Our room was very large with a lovely bathroom with a small provision for skylight. If you like the typical European breakfast, then it is, free, delicious and ample. The hotel is located by the Museum of Natural History (Museaodi Storia Nautrale). The hotel is also situated in a calm area away from the busy parts of the city but still in comfortable walking distance to everything. \\
\hline
4 &
The service was excellent. The staff was very pleasant and helpful. The rooms were very clean. The room was quite big, the bed was comfy, and it had a kitchenette with a good size fridge. The hotel is in walking distance to most major attractions and the public transportation is easily accessible. The hotel was very nice, and looks like they are doing renovations. The rooms need it. It’s not a high-end hotel, but \hl{it’s great value for the price.} &
The hotel, which is within walking distance to most major attractions, looked dated, like it needed renovations done. Aside from that, the staff was very friendly and helpful, going out of their way to please and assist. No room service, though. The rooms are quite clean and spacious, with comfy beds and a breadth of amenities such as a kitchenette and free internet &
Service at the front desk is very friendly and even arranges taxi service; overall just helpful. Everything was real clean, bright, and maintained daily. The rooms were large, with comfy beds, but the furniture was dated. The rest of the hotel could use renovations, it was all dated. The hotel is walking distance to major attractions and public transportation. \\
\hline
5 &
Staff were incredibly friendly which made our stay more comfortable. Room was a clean and tidy suite. I was plwasantly surprised by the room. The size was ok and it was small, there was a small kitchen, living room and separate bedroom. It also had big comfortable beds. Room service from the adjoining Moxie’s Grill has good food. The breakfast was very good, too, and can be taken up to a leisurely 2pm at the weekend. The hotel is also in a great location with all major downtown attractions within reasonable walking distance &
The hotel is in a great location with all major downtown attractions within reasonable walking distance. The relatively small rooms are comfortable, clean, and convenient, with a kitchen and washer/dryer at your service. The staff on reception were always friendly and helpful, and informative with any question you have to ask them. Room service from the adjoining Moxie’s Grill has good food, fast delivery and friendly service. &
The hotel staff was friendly, helpful, and informative. The housekeeping service was excellent and fast. The rooms and hotel are very clean, but people dislike that there are no suite cleaning services (no fresh towels/linens). Although small, rooms include a small kitchen, living room, and bedroom. The bed is big and comfy. There’s a good view from the balcony. Room service, breakfast, and Moxie’s Grill next door all have good food. The hotel is in a great location; all major downtown attractions are within walking distance. The \hl{price is good value}, compared to other hotels in Vancouver. \\
\hline
6 &
The location is excellent, right next to Placa Catalunia. The rooms are nice and clean and big, with great amenities and comfortable beds to satisfy. Every member of staff, from reception to bar staff were extremely professional and helpful, and the breakfast was good, with a sufficient range of offers. &
Tha hotel staff was delightful. They were always willing to help answer my questions or provide recommendations. The hotel, rooms, and bathrooms and clean and well maintained. The room was a good size, quite, and had very comfortable beds. Breakfast was good, with a sufficient range of offers. The location was excellent. It was close to everything, including shopping and the Placa Catalunia. \hl{The value and reasonable quality makes it hard to beat}. &
Reception, bar, and housekeeping staff were all delightful and helpful. The hotel was comfortably clean and fresh. The rooms were a good size, with very comfortable beds and great amenities. The hotel is located next to Placa Catalunia, and close enough to get a great shopping and cultural experience. \\
\hline
7 &
The hotel is characterized by a beautiful interior design, and it’s located close to everything, yet slightly out of the crush of tourist activity. The staff are wonderful and accommodating, and are ready to make all sorts of special arrangements for their guests upon request. The beautifully design and furnished room was modern and quite chic and very clean, as is the bathroom and shower room. The restaurant serves excellent full breakfasts, and their lunch and dinner menus are good too. &
The staff was very helpful with all of our questions, since it was our first time in Venice. It was very clean and for the stay. The rooms are large, very comfortable, with a beautiful “deco-design” furniture, and a great bathroom (jacuzzi...) with fine bathproducts. The restaurant and bar is small but very good quality - modern Venetian food and excellent breakfasts. Excellent location, close to everything, yet slightly out of the crush of the tourist hubbub. The rooms are characterized by a beautiful interior design, as all as the rest of the Hotel (lobby, restaurant, main floor). &
The hotel staff was knowledgeable, efficient, and always available to assist or give recommendations. They make many time-consuming special arrangements and reservations for guests. The hotel, rooms, and bathrooms were impeccably clean. The rooms are large, with beautiful deco furniture, comfortable beds, and quality linens. The bathrooms are small, but have great amenities, decor, and a jacuzzi. The restaurant and bar is small but great quality, with modern Venetian food and excellent breakfasts. Excellent location in a lovely neighborhood, close to everything, but away from crowds. The hotel interior design is beautiful. It’s not cheap, but it’s a \hl{great value}. \\
\hline
8 &
It’s a very nice hotel in a great location. The entire staff, from the doorman to the check-in, were very helpful and friendly. The rooms are very clean and generally well-kept by the staff, and they feature refrigerators and comfortable beds to sleep in. Restaurant in hotel is good for breakfast but best avoided for lunch and dinner for its lackluster menu with inconsistent quality. &
The entire hotel staff was very nice and helpful. The room was immaculately clean, as was the hotel. The rooms although small are beautifully designed, comfortably and stylishly equipped with almost everything, except perhaps a fridge. Beds and pillows were very comfortable. Restaurant in hotel is good for breakfast, but lunch and dinner have a lackluster menu with inconsistent quality. It’s a \hl{great value} for the wonderful location - right in the middle of everything. It’s much better than many of the chain hotels in the area. &
The entire hotel staff was very nice and helpful. The room was immaculately clean and well maintained. The rooms although small are beautifully designed, comfortably and stylishly equipped with just about everything you could want, with comfortable beds. The restaurant in hotel is good, too. The best part was the location - right in the middle of everything. \\
\hline
\caption{Qualitative examples of product summaries for eight hotels from the SPACE dataset where the ``Value for Money'' theme is explicitly discussed (e.g., ``good value'', ``reasonably priced''). These examples demonstrate that ``Value for Money'' is a salient and frequently recurring theme in user feedback, underscoring the limitation of the standard SPACE schema which excludes this dimension from evaluation.}
\label{tab:value_for_money_summaries_space}
\end{longtable}
\end{center}
\section{Theme-Constrained Opinion Extraction}

\subsection{SPACE: Theme Refinement and Consolidation}
\label{sec: space_theme_ref_and_consol}

This section details the stepwise refinement process applied to the raw themes extracted from the SPACE dataset. Starting with an initial set of 569 raw themes, we applied a three-stage filtering pipeline to derive a finalized, high-quality schema.

\subsubsection{Frequency based filtering}

In the initial unconstrained generation phase, the model identified a long tail of 569 distinct themes. To prioritize relevance, we analyzed the frequency distribution of these themes. Table~\ref{tab:category_counts} presents the top 22 most frequent themes.

We observed significant semantic redundancy in the raw output. For instance, synonymous terms such as ``view'' and ``views'' appeared as distinct categories. This fragmentation underscores the necessity of the subsequent semantic deduplication step to consolidate variations and improve aspect coverage.

\begin{table}[h!]
\centering
\begin{tabular}{l r}
\hline
\textbf{Category} & \textbf{Count} \\
\hline
rooms & 9820 \\
service & 5999 \\
location & 4534 \\
overall\_experience & 4176 \\
food & 3244 \\
amenities & 1847 \\
cleanliness & 1484 \\
value & 442 \\
noise & 339 \\
parking & 293 \\
internet & 245 \\
facilities & 173 \\
transportation & 160 \\
recommendation & 158 \\
pricing & 124 \\
decor & 111 \\
pool & 109 \\
bathroom & 108 \\
breakfast & 103 \\
view & 97 \\
safety & 97 \\
views & 85 \\
\hline
\end{tabular}
\caption{Frequency distribution of the top 22 raw themes identified in the SPACE dataset during the initial unconstrained generation phase (Section~\ref{subsubsec:unconstrained_theme_gen}). These counts represent the unfiltered output before refinement and semantic deduplication.}
\label{tab:category_counts}
\end{table}


\subsubsection{Semantic Deduplication}

In this step, we computed embeddings for theme names and merged semantically equivalent terms based on cosine similarity. Table~\ref{tab:category_counts_after_sem_dedup} illustrates the impact of this consolidation. Notably, ``view'' and ``views'' were merged into a single category, resulting in an aggregated count of 182. Similarly, the count for ``pricing'' increased (from 124 to 183) as it absorbed related terms, ensuring that minor lexical variations did not dilute the prominence of key aspects.

\begin{table}[h!]
\centering
\begin{tabular}{l r r}
\hline
\textbf{Final Theme} & \textbf{Original Count} & \textbf{Updated Count} \\
\hline
rooms & 9820 & 9820 \\
service & 5999 & 5999 \\
location & 4534 & 4534 \\
overall\_experience & 4176 & 4176 \\
food & 3244 & 3244 \\
amenities & 1847 & 1847 \\
cleanliness & 1484 & 1484 \\
value & 442 & 442 \\
noise & 339 & 339 \\
parking & 293 & 293 \\
internet & 245 & 245 \\
pricing & 124 & 183 \\
view & 97 & 182 \\
facilities & 173 & 173 \\
transportation & 160 & 160 \\
recommendation & 158 & 158 \\
decor & 111 & 111 \\
pool & 109 & 109 \\
bathroom & 108 & 108 \\
breakfast & 103 & 103 \\
safety & 97 & 97 \\
\hline
\end{tabular}
\caption{Frequency distribution of the top 22 raw themes identified in the SPACE dataset during the initial ``Unconstrained Theme Generation'' phase~\ref{subsubsec:unconstrained_theme_gen}. The counts reflect the raw output before and after semantic deduplication, highlighting the importance of semantic deduplication}
\label{tab:category_counts_after_sem_dedup}
\end{table}

\subsubsection{Human Review}

The final stage involved human-in-the-loop validation to incorporate domain expertise and business logic. This step was crucial for resolving ambiguities and deciding the optimal level of granularity for the final schema.
We compared the top themes identified by MOSAIC against the fixed schema used in the original SPACE dataset. Table~\ref{tab:additional_top6_themes} highlights frequent themes in our extraction—such as ``Value for Money'' and ``Bathroom'' -- that were often missed or subsumed in the original dataset.

\begin{table}[h!]
\centering
\begin{tabular}{l r}
\hline
\textbf{Theme} & \textbf{Count} \\
\hline
Value for Money & 25 \\
Bathroom        & 13 \\
Quietness       & 8 \\
Transportation  & 2 \\
\hline
\end{tabular}
\caption{Frequency of newly identified themes that appear in the top-six most frequent themes per product as per MOSAIC, but are absent from the original SPACE dataset.}
\label{tab:additional_top6_themes}
\end{table}

The following editorial decisions were made to finalize the theme list:

\begin{itemize} 
    \item \textbf{Exclusion of Overall Experience:} The theme overall\_experience was removed as it largely duplicates the aggregate review rating and does not provide specific, actionable insight. 
    \item \textbf{Granularity Adjustment (Building vs. Amenities):} The original SPACE dataset utilizes a broad ``Building'' category. Our extraction identified granular components such as amenities, parking, decor, and internet[1559]. We analyzed opinions within decor and found they spanned both room-specific and general hotel aesthetics[1560]. To balance detail with readability, we grouped these structural aspects under a consolidated ``Building \& Facilities'' theme. 
    \item \textbf{Bathroom:} Given its high frequency and distinct importance to user satisfaction, we retained bathroom as a standalone theme rather than merging it into rooms. 
    \item \textbf{Food \& Beverage:} The extracted theme breakfast was merged into the broader food category to maintain consistency.
\end{itemize}

In contrast to the 6 static themes predefined in the SPACE dataset, our analysis indicates that a comprehensive evaluation of this domain requires a broader and more granular schema. Consequently, we established a finalized set of \textbf{10 distinct themes} -- Rooms, Bathroom, Staff \& Service, Location, Cleanliness, Food, Transportation, Value for Money, Building \& Facilities, and Quietness —which better captures the diversity of user opinions found in the reviews.


\subsection{Impact of Base Model on Extraction Granularity}
\begin{table}[h!]
\centering
\small
\begin{tabular}{llcc}
\hline
\textbf{Dataset} & 
\textbf{Model} & 
\textbf{Avg. Opinions / Theme} & \textbf{Avg. Themes / Product} \\
\hline
SPACE   & GPT-4o            & 1411 & 10.88 \\
SPACE   & GPT-4.1           & 1687 & 10.92 \\
SPACE   & LLaMA 3.1 70B     & 1349 & 10.88 \\
\hline
PeerSum & GPT-4o            & 738  & 5.76  \\
PeerSum & GPT-4.1           & 1107 & 5.70  \\
PeerSum & LLaMA 3.1 70B     & 621  & 5.79  \\
\hline
\end{tabular}
\caption{Dataset statistics across different models, showing the average number of opinions per theme and the average number of themes per product using section~\ref{subsubsec:unconstrained_theme_gen}}
\label{tab:dataset_stats}
\end{table}

Table~\ref{tab:dataset_stats} illustrates how the choice of base model impacts the granularity of the extraction phase across both SPACE and PeerSum datasets. We observe that GPT-4.1 consistently extracts a higher density of information, identifying approximately 25
\section{Prompts}
\label{section:prompts_appendix}
\subsection{Prompts for Unconstrained Theme Generation}
\label{subsubsec:prompts_unconstrained_theme_gen}

This section presents the prompts used for unconstrained theme generation (Section~\ref{subsubsec:unconstrained_theme_gen}) for both PeerSum and SPACE.

    \begin{tcolorbox}[
        enhanced, breakable, 
        colback=black!5,
        colframe=PromptBlue,
        colbacktitle=PromptBlue,
        coltitle=white,
        fonttitle=\bfseries\small,
        title={\centering Unconstrained Theme Generation: SPACE},
        arc=2mm,
        boxrule=1.5pt,
        left=10pt, right=10pt, top=3pt, bottom=3pt 
    ]
        \small 
        \setlength{\parskip}{0.3em} 

        Analyze the hotel-related review text and extract key aspect terms with a strong focus on accuracy, context-awareness, and relevance to the hotel stay.

        \textbf{Read the Review Text as a Whole}
        \begin{itemize}
            \item Understand the full meaning of the review before extracting aspects.
            \item Consider nearby context to avoid misattributing sentiments.
            \item Avoid extracting aspects or opinions from sentence fragments; always evaluate the full sentence for meaning.
        \end{itemize}

        \textbf{Identify Aspect Terms}
        \begin{itemize}
            \item Extract only aspects directly related to the guest’s hotel experience (e.g., clean room, noisy AC, friendly staff).
            \item Do not classify statements as advice unless they provide explicit guidance to other travelers (e.g., booking suggestions or preparation tips).
            \item Avoid aspects related to other tours, general travel conditions, or external comparisons.
            \item If the same aspect repeats (e.g., ``clean room'' / ``very clean bathroom''), group them under a single aspect.
        \end{itemize}

        \textbf{Determine Opinion}
        \begin{itemize}
            \item Extract opinions only when they describe the guest’s experience with the hotel or provide actionable advice.
            \item Do not include comparisons to hotels not experienced by the reviewer.
            \item Opinions must be complete, clearly expressed, and accurately reflect sentiment.
        \end{itemize}

        \textbf{Assess Sentiment}
        \begin{itemize}
            \item Analyze complete sentences or paragraphs rather than isolated words.
            \item Classify sentiment strictly as positive, negative, or neutral.
            \item Mark aspects as negative only when dissatisfaction is explicitly expressed.
            \item Treat reflective or comparative statements as neutral.
        \end{itemize}

        \textbf{Classify Theme}
        \begin{itemize}
            \item Assign an appropriate theme to each aspect term (e.g., food, location, cleanliness).
        \end{itemize}

        \textbf{Output Format}
        \begin{itemize}
            \item Return a single JSON object.
            \item Include only themes with at least one aspect--opinion--sentiment triple.
            \item For a single aspect--opinion pair, return one object with keys: \texttt{aspect}, \texttt{opinion}, \texttt{sentiment}.
            \item For multiple pairs under the same theme, return an array of such objects.
            \item Do not return arrays for themes with only one aspect.
        \end{itemize}

        \textbf{Example 1} \\
        \textbf{Input:} <input\_text> \\
        \textbf{Output:} <output\_text> \\[0.5em]
        
        \textbf{Example 2} \\
        \textbf{Input:} <input\_text> \\
        \textbf{Output:} <output\_text> 

    \end{tcolorbox}
    \begin{tcolorbox}[
        colback=black!5,
        colframe=PromptGreen,
        colbacktitle=PromptGreen,
        coltitle=white,
        fonttitle=\bfseries\small,
        title={\centering Unconstrained Theme Generation: PeerSum},
        arc=2mm,
        boxrule=1.5pt,
        left=10pt, right=10pt, top=5pt, bottom=5pt 
    ]
        \small 
        \setlength{\parskip}{0.3em} 

        Analyze the peer review text of an academic manuscript and extract key aspect terms with a strong focus on accuracy, context-awareness, and relevance to scholarly evaluation criteria.

        \textbf{Read the Review Text as a Whole}
        \begin{itemize}
            \item Understand the full meaning of the review before extracting aspects.
            \item Consider nearby context to avoid misattributing sentiments.
            \item Avoid extracting aspects or opinions from sentence fragments; always evaluate the full sentence for meaning.
        \end{itemize}

        \textbf{Identify Aspect Terms}
        \begin{itemize}
            \item Extract only aspects that evaluate the manuscript itself.
            \item If the same scholarly aspect repeats (e.g., ``writing clarity'' / ``presentation is clear''), group them under a single academic aspect such as clarity.
        \end{itemize}

        \textbf{Determine Opinion}
        \begin{itemize}
            \item Extract opinions only when they describe the reviewer’s evaluation of the manuscript or provide actionable scholarly feedback (e.g., suggested improvements, concerns, or endorsements).
            \item Do not include comparisons to other papers, venues, or prior work unless explicitly evaluated in the review.
            \item Opinions must be complete, clear, tied to an aspect, and accurately represent the sentiment.
        \end{itemize}

        \textbf{Assess Sentiment}
        \begin{itemize}
            \item Analyze complete sentences and paragraphs rather than isolated words.
            \item Classify sentiment strictly as positive, negative, or neutral.
            \item Mark aspects as negative only if clear criticism, concern, or rejection-oriented feedback is expressed.
            \item Treat cautious or hedged language (e.g., ``may benefit from'', ``could be clearer'') as neutral unless clearly critical.
        \end{itemize}

        \textbf{Classify Theme}
        \begin{itemize}
            \item Assign an appropriate academic review theme (e.g., clarity, novelty) to each aspect term.
        \end{itemize}

        \textbf{Output Format}
        \begin{itemize}
            \item Return a single JSON object.
            \item Include only themes with at least one aspect--opinion--sentiment triple.
            \item If a theme contains one aspect--opinion pair, return a single object with keys: \texttt{aspect}, \texttt{opinion}, \texttt{sentiment}.
            \item If a theme contains multiple distinct pairs, return an array of such objects.
            \item Do not return arrays for themes with only one aspect.
        \end{itemize}

        \textbf{Example 1} \\
        \textbf{Input:} <input\_text> \\
        \textbf{Output:} <output\_text> \\[0.5em]
        
        \textbf{Example 2} \\
        \textbf{Input:} <input\_text> \\
        \textbf{Output:} <output\_text> 

    \end{tcolorbox}

\subsection{Prompts for Theme-Constrained Opinion Extraction - Recall Maximization}
\label{subsubsec:prompts_theme_constrained_opinion_extraction}

    \begin{tcolorbox}[
        colback=black!5,
        colframe=PromptBlue,
        colbacktitle=PromptBlue,
        coltitle=white,
        fonttitle=\bfseries\small,
        title={\centering Theme-Constrained Opinion Extraction: SPACE},
        arc=2mm,
        boxrule=1.5pt,
        left=10pt, right=10pt, top=5pt, bottom=5pt, breakable
    ]
        \small
        \setlength{\parskip}{0.3em}

You are a review analysis assistant and your job is to extract relevant aspect-opinion pairs from user reviews of hotel stays and classify them into the correct themes. You MUST only use the exact, case-sensitive theme names provided in the list below as keys in your final JSON output. DO NOT invent, alter, or create new themes. Every key in your output JSON must be one of the keys defined here. If no aspect from the review fits a theme, the value for that key MUST be null.

- \textbf{Rooms}: Captures feedback on the guest's private room, focusing on its comfort, features, and physical attributes. This includes the bed, pillows, linens, room size, furniture, and in-room decor. It also covers the functionality of in-room amenities like the TV, minibar, fridge, coffee maker, and air conditioning (AC). Opinions on the view from the room and the balcony belong here but opinions on views from public areas belong in Building \& Facilities theme. Excludes: All opinions on noise including AC noise which belong to quietness, cleanliness (which belong to cleanliness), or the bathroom (which is a separate theme).

- \textbf{Bathroom}: Captures all feedback specifically about the in-room bathroom. This includes its size, layout, and the functionality of its fixtures: shower, bathtub, water pressure, hot water availability, sink, and toilet. It also includes the quality and availability of towels, bathrobes, and provided toiletries (e.g., shampoo, soap). Excludes: Opinions on bathroom cleanliness (which belong to the cleanliness theme).

- \textbf{Staff \& Service}: Captures feedback on hotel personnel, focusing on their attitude and competence. This includes traits like friendliness, helpfulness, professionalism, and attentiveness. It covers all departments: front desk, check-in/out, concierge, valet, bellhop, and housekeeping. This theme also includes the speed and politeness of room service. Excludes: Feedback on the quality or taste of room service food (which belongs to the food theme) and the quality/result of the cleaning (which belongs to cleanliness).

- \textbf{Location}: Captures feedback on the hotel's geographic position and convenience relative to external points. This primarily includes walkability and proximity to tourist attractions, restaurants, shops, and public transit hubs (e.g., metro/subway stations, bus stops, train stations, and the airport). It also covers the neighborhood's character, safety, and atmosphere. Excludes: Opinions on hotel-operated shuttle services (which belong to transportation).

- \textbf{Cleanliness}: Captures all feedback on hygiene, tidiness, and the result of the cleaning service. This includes the cleanliness of private spaces (e.g., room, bathroom) and public areas (e.g., lobby, pool). This theme explicitly covers the performance and thoroughness of housekeeping (e.g., "room was cleaned well," "they missed a spot").

- \textbf{Food}: Captures feedback on the quality, taste, variety, and presentation of all food and beverages provided by the hotel itself. This includes the restaurant, bar, breakfast (buffet, continental, or complimentary), and the quality/taste of room service food. Opinions on in-room items like welcome snacks or bottled water also belong here. Excludes: Opinions on food prices (which belong to value for money) or restaurant decor (which belongs to building and facilities) or any opinions about external or next-door restaurants not operated by the hotel.

- \textbf{Transportation}: Captures feedback only on transportation services actively provided or arranged by the hotel. This theme is strictly for hotel-operated services like the free shuttle, airport shuttle, or courtesy car. It covers their reliability, frequency, cost, and convenience. Excludes: General taxi availability or public transit access (which belong to location).

- \textbf{Value for Money}: Captures all feedback related to price, cost, and perceived value. This includes opinions on the overall room rate (e.g., "expensive," "good value," "worth the price") as well as the cost of specific billable items like parking, internet access, breakfast, or minibar items.

- \textbf{Building \& Facilities}: Captures feedback on the physical property, shared amenities, and overall aesthetics. This includes public spaces and décor (lobby, lounges, elevators, grounds, patios, architecture, atmosphere), recreational facilities (pool, gym, fitness center, spa, hot tub), business and logistical spaces (parking, business center, laundry, conference/event rooms), property-wide utilities (Wi-Fi, internet, public computers), and views from public areas (e.g., rooftop bar).

- \textbf{Quietness}: Captures all feedback about noise levels or the lack of noise. This includes noise from external sources (street, traffic, construction, nightlife) and internal sources (hallways, neighbors, elevators, in-room AC, plumbing). This covers both positive (quiet, peaceful) and negative (loud, noisy) opinions.

- \textbf{Other}: Miscellaneous feedback that doesn’t fall into the other themes.

Think step by step and follow the instructions below:

\textbf{1. Read the Review Text as a Whole}
\begin{itemize}
    \item Understand the full meaning of the review before extracting aspects.
    \item Use surrounding context to ensure correct attribution of opinions.
\end{itemize}

\textbf{2. Identify Aspect Terms}
\begin{itemize}
    \item Extract terms or phrases (“aspect”) that clearly refer to a hotel-related entity, concept, or theme.
    \item The "aspect" field must concisely state what is being directly evaluated (e.g., "staff friendliness", "bed comfort", "pool cleanliness").
\end{itemize}

\textbf{3. Determine Opinion}
\begin{itemize}
    \item Extract the exact phrase(s) from the review that express sentiment about the aspect only.
    \item Double check if the right opinion is extracted from the review.
    \item If multiple distinct opinions are identified for a single theme, extract each as a separate aspect-opinion pair.
    \item If a single opinion phrase applies to multiple themes, extract and classify under each relevant theme.
    \item Do not include unrelated commentary or context about other topics.
\end{itemize}

\textbf{4. Assess Sentiment}
\begin{itemize}
    \item Sentiment must directly match the extracted opinion and aspect.
    \item Classify sentiment as positive, negative, or neutral only.
    \item Mark aspects as negative only if dissatisfaction is expressed.
    \item Classify statements as neutral when they convey facts, general observations, reflective thoughts, curiosity, or broad comparisons without clear positive/negative sentiment.
\end{itemize}

\textbf{5. Classify Theme}
\begin{itemize}
    \item Assign the correct theme based only on the definitions provided.
    \item Use the exact category name (case-sensitive). Do not alter the casing.
\end{itemize}

\textbf{Output Format:} Your response MUST be a single JSON object using the exact theme names defined above. Each key’s value must be either:
\begin{itemize}
    \item An object with fields: "aspect", "opinion", "sentiment", if there is exactly one aspect-opinion pair.
    \item An array of objects with fields: "aspect", "opinion", "sentiment", if there are multiple pairs.
    \item null, if the theme is not mentioned in the review.
\end{itemize}

\textbf{Example 1} \\
\textbf{Input:} <input\_text> \\
\textbf{Output:} <output\_text> \\[0.5em]

\textbf{Example 2} \\
\textbf{Input:} <input\_text> \\
\textbf{Output:} <output\_text>

    \end{tcolorbox}

    \begin{tcolorbox}[
        colback=black!5,
        colframe=PromptGreen,
        colbacktitle=PromptGreen,
        coltitle=white,
        fonttitle=\bfseries\small,
        title={\centering Theme-Constrained Opinion Extraction: PeerSum},
        arc=2mm,
        boxrule=1.5pt,
        left=10pt, right=10pt, top=5pt, bottom=5pt, breakable 
    ]
        \small 
        \setlength{\parskip}{0.3em} 

        You are an academic peer-review and discussion analysis assistant. Your job is to extract
aspect--opinion pairs from all evaluative and argumentative content in source documents
(including reviewer feedback, author responses, meta-reviews, and threaded discussions) and
classify them into the correct themes.

You MUST only use the exact, case-sensitive theme names provided in the list below as keys in
your final JSON output. DO NOT invent, alter, or create new themes. Every key in your output
JSON must be one of the keys defined here. If no aspect from the review fits a theme, the value
for that key MUST be null.

\begin{itemize}
    \item \textbf{Advancement}: Opinions about the significance and concrete contributions of the work in advancing the research field. This includes empirical improvements over strong baselines, new capabilities, efficiency gains, theoretical or practical impact, and how much the work helps future research or applications. Advancement may be recognized even when novelty is incremental, as long as the work delivers meaningful progress. Do not use Advancement when a comment focuses only on how original or derivative the idea is; those belong to Novelty.
    \item \textbf{Novelty}: Opinions about the originality and innovation of the work. This includes whether the paper introduces genuinely new ideas, formulations, methods, architectures, datasets, or theoretical insights, and how distinct it is from prior or state-of-the-art work. It covers phrases such as ``novel contribution,'' ``incremental,'' or ``already known.'' Do not use Novelty when a comment mainly evaluates the importance or practical impact of the work rather than how new it is; those belong to Advancement.
    \item \textbf{Clarity}: Opinions about the writing quality, readability, and presentation of the paper. This includes organization and flow, completeness and accessibility of explanations, quality and interpretability of figures and tables, notation and terminology, and mechanical issues such as grammar or formatting. Use Clarity for cases where ideas might be sound but are hard to follow, and for reviewer confusion or requests for clarification. Do not use Clarity when the main point is that methods, experiments, or arguments are incorrect or unsupported; those belong to Soundness.
    \item \textbf{Compliance}: Opinions about the paper’s adherence to venue policies, ethical standards, and procedural guidelines. This includes fit to the venue’s scope, anonymity violations, missing ethics approvals, data/privacy or consent issues, non-compliance in user studies, citation or formatting requirements, and other policy-level concerns. Do not use Compliance for the scientific quality of methods or experiments (Soundness) or for the importance of the contribution (Advancement).
    \item \textbf{Soundness}: Opinions about the technical correctness, methodological validity, and strength of evidence. This includes the appropriateness of the methodology, experimental design and baselines, correctness and rigor of theory or proofs, quality of analysis, reliability of results, and whether claims are adequately supported. Use Soundness when reviewers say things like ``experiments are flawed,'' ``evidence is convincing,'' ``claims are oversold,'' or ``proofs are not rigorous.'' Do not use Soundness when the comment is mainly about readability or organization; those belong to Clarity.
    \item \textbf{Other}: Miscellaneous feedback that does not fall into the other themes.
\end{itemize}

Think step by step and follow the instructions below:

\textbf{1. Read the Review Text as a Whole}
\begin{itemize}
    \item Understand the context and identify whether the text is a critique (Reviewer) or a defense/rebuttal (Author).
    \item Use surrounding context to ensure correct attribution of opinions.
\end{itemize}

\textbf{2. Identify Aspect Terms}
\begin{itemize}
    \item Extract terms or phrases (``aspect'') that clearly refer to a peer review of an academic manuscript.
    \item The \texttt{aspect} field must concisely state what is being directly evaluated (e.g., ``writing clarity,'' ``conference suitability,'' ``experimental design'').
\end{itemize}

\textbf{3. Determine Opinion}
\begin{itemize}
    \item Extract the exact phrase(s) from the text that express sentiment about the aspect only.
    \item For reviewers: extract phrases expressing judgment, praise, or criticism.
    \item For authors: extract phrases that state a claim, counter-argument, clarification, or defense.
    \item If multiple distinct opinions are identified for a single theme, extract each as a separate aspect--opinion pair.
    \item If a single opinion phrase genuinely applies to multiple themes, extract and classify it under each relevant theme.
    \item Do not include unrelated commentary or background context.
\end{itemize}

\textbf{4. Assess Sentiment}
\begin{itemize}
    \item For reviewers:
    \begin{itemize}
        \item Positive: Praise or satisfaction.
        \item Negative: Criticism or dissatisfaction.
        \item Neutral: Factual observations or questions.
    \end{itemize}
    \item For authors (rebuttals):
    \begin{itemize}
        \item Positive: Defending validity, asserting contributions, or refuting a flaw.
        \item Negative: Conceding a flaw or acknowledging a limitation.
        \item Neutral: Pure clarification or explanation without judgment.
    \end{itemize}
\end{itemize}

\textbf{5. Classify Theme}
\begin{itemize}
    \item Assign the correct theme strictly based on the provided definitions.
    \item Use the exact, case-sensitive theme name.
\end{itemize}

\textbf{Output Format}
\begin{itemize}
    \item Return a single JSON object.
    \item The top-level keys MUST be the exact theme names defined above.
    \item Each value must be:
    \begin{itemize}
        \item An object with fields \texttt{aspect}, \texttt{opinion}, \texttt{sentiment} if there is exactly one pair.
        \item An array of such objects if multiple pairs exist.
        \item \texttt{null} if the theme is not mentioned.
    \end{itemize}
\end{itemize}

        \textbf{Example 1} \\
        \textbf{Input:} <input\_text> \\
        \textbf{Output:} <output\_text> \\[0.5em]
        
        \textbf{Example 2} \\
        \textbf{Input:} <input\_text> \\
        \textbf{Output:} <output\_text> 

    \end{tcolorbox}

\subsection{Prompts for Theme-Constrained Opinion Extraction - Precision Refinement}
\label{subsubsec:prompts_theme_constrained_opinion_extraction_pr}
~
    \begin{tcolorbox}[
        enhanced jigsaw, breakable,
        colback=black!5,
        colframe=PromptBlue,
        colbacktitle=PromptBlue,
        coltitle=white,
        fonttitle=\bfseries\small,
        title={Precision Refinement: SPACE},
        center title,
        arc=2mm,
        boxrule=1.5pt,
        left=10pt, right=10pt, top=3pt, bottom=3pt,
        before upper={\small\setlength{\parskip}{0.3em}}
    ]
        {\small
        \setlength{\parskip}{0.3em}
You are an expert in reviewing the output of another AI model.

You will be provided with:
\begin{itemize}
    \item A \textbf{user review} describing a hotel stay or experience
    \item A \textbf{theme label} along with its \textbf{definition}
    \item A \textbf{model output} consisting of an aspect--opinion pair and sentiment assigned to that theme
\end{itemize}

Your task is to determine whether the theme label used in the model output is appropriate.

\textbf{Decision Criteria}

Respond with \textbf{``Yes''} only if:
\begin{itemize}
    \item The quoted opinion in the model output clearly aligns with the theme definition
    \item All inclusion criteria are satisfied and no exclusion criteria are violated
    \item The user review is used only to clarify the meaning or tone of the quoted opinion
\end{itemize}

Respond with \textbf{``No''} if:
\begin{itemize}
    \item The quoted opinion does not fit the theme definition, even with context
    \item Any exclusion criteria are violated
    \item The theme appears justified only due to other, unrelated opinions in the review
\end{itemize}

\textbf{Additional Instructions}
\begin{itemize}
    \item Base your decision strictly on the quoted opinion; the review is only supporting context
    \item Do not substitute or override the quoted opinion with stronger phrases elsewhere in the review
    \item Do not infer relevance beyond what is explicitly expressed
\end{itemize}

Respond with \textbf{``Yes''} or \textbf{``No''}, followed by a concise justification.

        \textbf{Example 1} 
        
        \textbf{Input:} <input\_text> 
        
        \textbf{Output:} <output\_text> 
        
        \vspace{0.5em}
        \textbf{Example 2} 
        
        \textbf{Input:} <input\_text> 
        
        \textbf{Output:} <output\_text> 
        
        \vspace{0.5em}
        \textbf{Example 3} 
        
        \textbf{Input:} <input\_text> 
        
        \textbf{Output:} <output\_text>
    }
    \end{tcolorbox}
    
    \begin{tcolorbox}[
        enhanced, breakable, before skip=0pt, after skip=0pt,
        colback=black!5,
        colframe=PromptBlue,
        colbacktitle=PromptBlue,
        coltitle=white,
        fonttitle=\bfseries\small,
        title={\centering Precision Refinement: PeerSum},
        arc=2mm,
        boxrule=1.5pt,
        left=10pt, right=10pt, top=5pt, bottom=5pt, 
    ]
        {\small
        \setlength{\parskip}{0.3em}

You are an expert reviewer evaluating the output of another AI model.

You will be provided with:
\begin{itemize}
    \item Reviewer feedback, author responses, meta-reviews, and threaded discussion from an academic peer-review process
    \item A \textbf{theme label} and its \textbf{definition}
    \item A \textbf{model output} consisting of a quoted opinion, its associated aspect, and sentiment
\end{itemize}

Your task is to determine whether the theme label assigned in the model output is appropriate.

\textbf{Decision Criteria}

Answer \textbf{``Yes''} only if:
\begin{itemize}
    \item The quoted opinion clearly fits the theme definition when interpreted using its immediate context
    \item The opinion satisfies all inclusion criteria and violates none of the exclusion criteria
    \item The surrounding input text is used only to clarify the meaning or tone of the quoted opinion
\end{itemize}

Answer \textbf{``No''} if:
\begin{itemize}
    \item The quoted opinion does not align with the theme definition, even with context
    \item Any exclusion criteria are violated
    \item The theme label appears justified only due to other, unrelated opinions in the review
\end{itemize}

\textbf{Additional Instructions}
\begin{itemize}
    \item Focus strictly on the quoted opinion from the model output
    \item Do not substitute or override the quoted opinion with other phrases from the review
    \item Do not infer relevance beyond what is explicitly expressed
\end{itemize}

\textbf{Output Format}

Respond with \textbf{``Yes''} or \textbf{``No''}, followed by a brief justification.

        \textbf{Example 1} \\
        \textbf{Input:} <input\_text> \\
        \textbf{Output:} <output\_text> \\[0.5em]
        \textbf{Example 2} \\
        \textbf{Input:} <input\_text> \\
        \textbf{Output:} <output\_text> 
    }
    \end{tcolorbox}

\subsubsection{Prompts for theme summary}
\label{subsubsec:theme_summary_prompt}

We use the prompt from \cite{li2025decomposed} for fair comparison.

    \begin{tcolorbox}[
        colback=black!5,
        colframe=PromptBlue,
        colbacktitle=PromptBlue,
        coltitle=white,
        fonttitle=\bfseries\small,
        title={\centering Theme Summary: SPACE},
        arc=2mm,
        boxrule=1.5pt,
        left=10pt, right=10pt, top=5pt, bottom=5pt, breakable
    ]
        \small
        \setlength{\parskip}{0.3em}

You are good at writing summaries for opinionated texts. You are given some opinionated text fragments, please write a concise summary for them.

\textbf{Output Format}

Respond with \textbf{valid JSON only} in the following structure:
\begin{verbatim}
{
  "summary": "<theme-level summary here>"
}
\end{verbatim}

    \end{tcolorbox}

    \begin{tcolorbox}[
        colback=black!5,
        colframe=PromptGreen,
        colbacktitle=PromptGreen,
        coltitle=white,
        fonttitle=\bfseries\small,
        title={\centering Theme Summary: PeerSum},
        arc=2mm,
        boxrule=1.5pt,
        left=10pt, right=10pt, top=5pt, bottom=5pt, 
    ]
        \small 
        \setlength{\parskip}{0.3em} 

        You are good at writing summaries for opinionated texts. You are given some opinionated text fragments, please write a concise summary for them.

\textbf{Output Format}

Respond with \textbf{valid JSON only} in the following structure:
\begin{verbatim}
{
  "summary": "<theme-level summary here>"
}
\end{verbatim}

    \end{tcolorbox}

\subsubsection{Prompts for Product Summary}
\label{subsubsec:product_summary_prompt}

We adopt the same prompt as \cite{li2025decomposed} to ensure a fair comparison.

    \begin{tcolorbox}[
        colback=black!5,
        colframe=PromptBlue,
        colbacktitle=PromptBlue,
        coltitle=white,
        fonttitle=\bfseries\small,
        title={\centering Product Summary: SPACE},
        arc=2mm,
        boxrule=1.5pt,
        left=10pt, right=10pt, top=5pt, bottom=5pt, breakable
    ]
        \small
        \setlength{\parskip}{0.3em}

You are good at understanding documents with hotel review opinions.
You are given business reviews covering different aspects of a hotel experience.

Your task is to write a concise, coherent, and natural meta-review that summarizes
the provided comments and covers all mentioned review aspects. The summary should
reflect the overall sentiment and key points without repeating individual reviews
verbatim or introducing new information.

\textbf{Output Format}

Respond with \textbf{valid JSON only} in the following structure:
\begin{verbatim}
{
  "summary": "<product-level summary here>"
}
\end{verbatim}

    \end{tcolorbox}

    \begin{tcolorbox}[
        colback=black!5,
        colframe=PromptGreen,
        colbacktitle=PromptGreen,
        coltitle=white,
        fonttitle=\bfseries\small,
        title={\centering Product Summary: PeerSum},
        arc=2mm,
        boxrule=1.5pt,
        left=10pt, right=10pt, top=5pt, bottom=5pt, breakable 
    ]
        \small 
        \setlength{\parskip}{0.3em} 

        You are good at understanding documents with scientific review opinions.
You are given reviewer comments covering different evaluative aspects of an
academic manuscript.

Your task is to write a concise, coherent, and natural meta-review that summarizes
the provided comments and covers all mentioned review aspects. The summary should
reflect the overall assessment and key concerns or strengths without repeating
individual comments verbatim or introducing new information.

\textbf{Output Format}

Respond with \textbf{valid JSON only} in the following structure:
\begin{verbatim}
{
  "summary": "<manuscript-level summary here>"
}
\end{verbatim}
        \textbf{Example 1} \\
        \textbf{Input:} <input\_text> \\
        \textbf{Output:} <output\_text> \\[0.5em]
        
        \textbf{Example 2} \\
        \textbf{Input:} <input\_text> \\
        \textbf{Output:} <output\_text> 

    \end{tcolorbox}

\subsubsection{Prompts for Importance of Opinion Clustering}
\label{subsubsec:opinion_redundancy_prompt}

    \begin{tcolorbox}[
        colback=black!5,
        colframe=PromptOrange,
        colbacktitle=PromptOrange,
        coltitle=white,
        fonttitle=\bfseries\small,
        title={\centering Generate Theme Summary: Opinion Clustering},
        arc=2mm,
        boxrule=1.5pt,
        left=10pt, right=10pt, top=5pt, bottom=5pt, breakable
    ]
        \small
        \setlength{\parskip}{0.3em}

You are analyzing customer reviews of tour experience products and activities.
Each review contains opinions related to specific themes such as Tour Guide,
Pickup Logistics, Food and Beverage, and related aspects of the experience.

Your task is to write a concise, coherent, and natural summary that captures
the overall sentiment and the most commonly mentioned points across reviews,
without repeating details or using reviewer phrases verbatim.

\textbf{Strict Guidelines}
\begin{itemize}
    \item Write a clear, focused, and concise summary between 35 and 50 words.
    \item Ensure all statements are directly supported by the source opinions;
    do not introduce assumptions or details not present in the text.
    \item Maintain a clear logical flow so the summary reads as a single cohesive idea.
    \item Focus only on the main recurring points; omit rare or incidental remarks.
\end{itemize}

\textbf{Process}
Before writing the final summary, follow these steps while strictly adhering
to the guidelines above:
\begin{enumerate}
    \item Identify the most important recurring points from the input reviews.
    \item Capture all key points while maintaining logical and coherent flow.
    \item Construct a concise summary between 35--50 words that follows all
    strict guidelines.
\end{enumerate}

\textbf{Output Format}

Respond with \textbf{valid JSON only} in the following structure:
\begin{verbatim}
{
  "summary": "<theme-level summary here>"
}
\end{verbatim}

    \end{tcolorbox}

    \begin{tcolorbox}[
        colback=black!5,
        colframe=PromptOrange,
        colbacktitle=PromptOrange,
        coltitle=white,
        fonttitle=\bfseries\small,
        title={\centering Coverage: Opinion Clustering},
        arc=2mm,
        boxrule=1.5pt,
        left=10pt, right=10pt, top=5pt, bottom=5pt, breakable
    ]
        \small
        \setlength{\parskip}{0.3em}

You are an expert evaluator. Your task is to determine which summary
covers more key points from a given input text.

You will be provided with:
\begin{itemize}
    \item An input text
    \item Two summaries (summary1 and summary2)
\end{itemize}

\textbf{Evaluation Rules}
\begin{itemize}
    \item Focus \textbf{only} on coverage of key information, facts, and attributes
    present in the input text.
    \item Ignore names, specific people, or entities mentioned in the input text;
    they are not important for this evaluation.
    \item Ignore grammar, writing style, tone, or phrasing.
    \item Identify which summary includes more distinct and relevant points
    mentioned in the input.
    \item Do \textbf{not} penalize missing names or minor details; evaluate only
    whether the main points are covered.
\end{itemize}

\textbf{Output Format}

Respond with \textbf{valid JSON only} in the following structure:
\begin{verbatim}
{
  "answer": "<1 | 2 | 3>",
  "reasoning": "<brief explanation of why this summary provides better or equal coverage>"
}
\end{verbatim}

        \textbf{Example 1} \\
        \textbf{Input:} <input\_text> \\
        \textbf{Output:} <output\_text> \\[0.5em]
        
        \textbf{Example 2} \\
        \textbf{Input:} <input\_text> \\
        \textbf{Output:} <output\_text> \\[0.5em]
        \textbf{Example 3} \\
        \textbf{Input:} <input\_text> \\
        \textbf{Output:} <output\_text>

    \end{tcolorbox}

    \begin{tcolorbox}[
        enhanced, breakable, 
        colback=black!5,
        colframe=PromptOrange,
        colbacktitle=PromptOrange,
        coltitle=white,
        fonttitle=\bfseries\small,
        title={\centering Faithfulness: Opinion Clustering},
        arc=2mm,
        boxrule=1.5pt,
        left=10pt, right=10pt, top=3pt, bottom=3pt
    ]
        \small
        \setlength{\parskip}{0.3em}

You are an expert evaluator. Your task is to determine which of two summaries
is more faithful to the given input text. Ignore missing details and do not
penalize a summary for leaving out information present in the input. Evaluate
only whether the claims made in the summaries are accurate with respect to the
input text.

\textbf{You will receive:}
\begin{itemize}
    \item An input text
    \item Two summaries (summary1 and summary2)
\end{itemize}

\textbf{Faithfulness Evaluation Rules}
\begin{enumerate}
    \item Extract factual claims from each summary.
    \item For each claim, determine whether it is:
    \begin{itemize}
        \item \textbf{Supported}: Explicitly stated or directly paraphrased
        from the input text.
        \item \textbf{Unsupported}: Not mentioned in the input text.
        \item \textbf{Contradicted}: Directly conflicts with the input text.
    \end{itemize}
    \item Do \textbf{not} penalize for missing details, writing style, grammar,
    or length. Evaluate only factual correctness relative to the input.
    \item Ignore missing details entirely; summaries do not need to cover all
    information present in the input text.
    \item Ignore opinions, interpretations, or generalizations unless they
    directly contradict the input text.
    \begin{itemize}
        \item Generalizations or paraphrases that accurately reflect the
        combined meaning of multiple input statements (e.g., collective
        references to multiple guides) should be treated as supported.
    \end{itemize}
    \item Compare the two summaries:
    \begin{itemize}
        \item The summary with fewer unsupported or contradicted claims is
        considered more faithful.
        \item If both summaries contain zero issues or the same number of
        issues, they are considered equally faithful.
    \end{itemize}
\end{enumerate}

\textbf{Output Format}

Respond with \textbf{valid JSON only} in the following structure:
\begin{verbatim}
{
  "answer": "<1 | 2 | 3>",
  "reasoning": "<brief explanation highlighting which claims in each summary are supported,
  unsupported, or contradicted, and why the chosen summary is more faithful>"
}
\end{verbatim}
        \textbf{Example 1} \\
        \textbf{Input:} <input\_text> \\
        \textbf{Output:} <output\_text> \\[0.5em]
        
        \textbf{Example 2} \\
        \textbf{Input:} <input\_text> \\
        \textbf{Output:} <output\_text> \\[0.5em]
        \textbf{Example 3} \\
        \textbf{Input:} <input\_text> \\
        \textbf{Output:} <output\_text>

    \end{tcolorbox}

\subsubsection{Prompts for SPACE deep dive}
\label{subsubsec:space_deep_dive_prompts}

    \begin{tcolorbox}[
        colback=black!5,
        colframe=PromptBlue,
        colbacktitle=PromptBlue,
        coltitle=white,
        fonttitle=\bfseries\small,
        title={\centering SPACE deep dive: Sentiment Score},
        arc=2mm,
        boxrule=1.5pt,
        left=10pt, right=10pt, top=5pt, bottom=5pt, breakable
    ]
        \small
        \setlength{\parskip}{0.3em}

You are an expert sentiment evaluator. Your task is to read a theme-level
summary extracted from hotel reviews and assign an overall sentiment score
between \textbf{0} and \textbf{100}, where:
\begin{itemize}
    \item \textbf{0} = Extremely negative sentiment
    \item \textbf{100} = Extremely positive sentiment
    \item \textbf{50} = Mixed or neutral sentiment
\end{itemize}

\textbf{Evaluation Rules}
\begin{itemize}
    \item Read the theme summary carefully and evaluate only the sentiment
    expressed in that summary.
    \item If the summary contains mostly positive sentiment with little or no
    negativity, assign a high score (80--100).
    \item If the summary contains mostly negative sentiment with little or no
    positivity, assign a low score (0--20).
    \item If the summary reflects a balanced or mixed sentiment, assign a
    mid-range score (30--70).
    \item Weigh strong or emphatic statements more heavily than mild or tentative
    ones.
    \item Ignore missing details or topics not mentioned in the summary.
\end{itemize}

\textbf{Output Format}

Respond with \textbf{valid JSON only} in the following structure:
\begin{verbatim}
{
  "score": <integer from 0 to 100>
}
\end{verbatim}

    \end{tcolorbox}

\subsubsection{G-eval}
\label{subsubsec:geval_prompts}

To ensure a fair comparison, we use the same prompt for G-eval as \cite{li2025decomposed}.

    \begin{tcolorbox}[
        colback=black!5,
        colframe=PromptBlue,
        colbacktitle=PromptBlue,
        coltitle=white,
        fonttitle=\bfseries\small,
        title={\centering G-eval: SPACE},
        arc=2mm,
        boxrule=1.5pt,
        left=10pt, right=10pt, top=5pt, bottom=5pt, breakable
    ]
        \small
        \setlength{\parskip}{0.3em}

Here are several review documents that contain opinions from different people about a hotel, along 
with a candidate summary of these reviews.

You are required to evaluate how accurately the given summary reflects the overall opinions for 
review aspects expressed in the original reviews.

Please read all opinions in the summary and calculate the percentage of faithful opinions that 
are clearly supported by the source review documents.

The percentage of faithful opinions (only output a decimal like 0.12, no other content)

    \end{tcolorbox}

    \begin{tcolorbox}[
        colback=black!5,
        colframe=PromptGreen,
        colbacktitle=PromptGreen,
        coltitle=white,
        fonttitle=\bfseries\small,
        title={\centering G-eval: PeerSum},
        arc=2mm,
        boxrule=1.5pt,
        left=10pt, right=10pt, top=5pt, bottom=5pt 
    ]
        \small 
        \setlength{\parskip}{0.3em} 

 Here are several review documents that contain opinions from different people about a scientific paper, along with a
candidate summary of these reviews.

You are required to evaluate how accurately the given summary reflects the overall opinions for review aspects expressed
in the original reviews.

Please read all opinions in the summary and calculate the percentage of faithful opinions that are clearly supported by
the source review documents.

The percentage of faithful opinions (only output a decimal like 0.12, no other content)
\end{tcolorbox}

\subsubsection{SPACE: theme identification}
\label{subsubsec:space_theme_identification}
Theme-identification prompts for the SPACE dataset are used in two settings: (i) to compute aspect coverage in Section~\ref{subsection: eval_metrics} and (ii) to compute theme coverage in Section~\ref{sec:space_data_deepdive}. We use the exact same prompt template as \cite{li2025decomposed}; the only modification is to the theme definitions, which are made more detailed to reduce overlap between closely related themes.

    \begin{tcolorbox}[
        colback=black!5,
        colframe=PromptPurple,
        colbacktitle=PromptPurple,
        coltitle=white,
        fonttitle=\bfseries\small,
        title={\centering Theme Identification: Building \& Facilities},
        arc=2mm,
        boxrule=1.5pt,
        left=10pt, right=10pt, top=5pt, bottom=5pt, breakable
    ]
        \small
        \setlength{\parskip}{0.3em}

You are good at understanding documents with hotel review opinions.
Below is a business review for a hotel, please extract fragments that are related to Building \& Facilities of the hotel.

Definition of Building \& Facilities:
Captures feedback on the physical property, shared amenities, and overall aesthetics. This includes public spaces and décor (lobby, lounges, elevators, grounds, patios, architecture, atmosphere), recreational facilities (pool, gym, fitness center, spa, hot tub), business and logistical spaces (parking, business center, laundry, conference/event rooms), property-wide utilities (Wi-Fi, internet, public computers), and views from public areas (e.g., rooftop bar).

Final extracted fragments (follow the format above in different lines and if no resulted fragments just output "No related fragments")
\end{tcolorbox}

    \begin{tcolorbox}[
        colback=black!5,
        colframe=PromptPurple,
        colbacktitle=PromptPurple,
        coltitle=white,
        fonttitle=\bfseries\small,
        title={\centering Theme Identification: Cleanliness},
        arc=2mm,
        boxrule=1.5pt,
        left=10pt, right=10pt, top=5pt, bottom=5pt, breakable
    ]
        \small
        \setlength{\parskip}{0.3em}

You are good at understanding documents with hotel review opinions.
Below is a business review for a hotel, please extract fragments that are related to Cleanliness of the hotel.

Definition of Cleanliness:
Captures all feedback on hygiene, tidiness, and the result of the cleaning service. 
This includes the cleanliness of private spaces (e.g., room, bathroom) and public areas (e.g., lobby, pool). 
This theme explicitly covers the performance and thoroughness of housekeeping (e.g., ``room was cleaned well,'' ``they missed a spot'').

Final extracted fragments (follow the format above in different lines and if no resulted fragments just output "No related fragments")
\end{tcolorbox}

    \begin{tcolorbox}[
        colback=black!5,
        colframe=PromptPurple,
        colbacktitle=PromptPurple,
        coltitle=white,
        fonttitle=\bfseries\small,
        title={\centering Theme Identification: Food},
        arc=2mm,
        boxrule=1.5pt,
        left=10pt, right=10pt, top=5pt, bottom=5pt, breakable
    ]
        \small
        \setlength{\parskip}{0.3em}

You are good at understanding documents with hotel review opinions.
Below is a business review for a hotel, please extract fragments that are related to Food of the hotel.

Definition of Food:
Captures feedback on the quality, taste, variety, and presentation of all food and beverages provided by the hotel itself. This includes the restaurant, bar, breakfast (buffet, continental, or complimentary), and the quality/taste of room service food. Opinions on in-room items like welcome snacks or bottled water also belong here. Excludes: Opinions on food prices (which belong to value for money) or restaurant decor (which belongs to building and facilities) or any opinions about external or next-door restaurants not operated by the hotel.

Final extracted fragments (follow the format above in different lines and if no resulted fragments just output "No related fragments")
\end{tcolorbox}

    \begin{tcolorbox}[
        colback=black!5,
        colframe=PromptPurple,
        colbacktitle=PromptPurple,
        coltitle=white,
        fonttitle=\bfseries\small,
        title={\centering Theme Identification: Location},
        arc=2mm,
        boxrule=1.5pt,
        left=10pt, right=10pt, top=5pt, bottom=5pt, breakable
    ]
        \small
        \setlength{\parskip}{0.3em}

You are good at understanding documents with hotel review opinions.
Below is a business review for a hotel, please extract fragments that are related to Location of the hotel.

Definition of Location:
Captures feedback on the hotel's geographic position and convenience relative to external points. This primarily includes walkability and proximity to tourist attractions, restaurants, shops, and public transit hubs (e.g., metro/subway stations, bus stops, train stations, and the airport). It also covers the neighborhood's character, safety, and atmosphere. Excludes: Opinions on hotel-operated shuttle services (which belong to transportation).

Final extracted fragments (follow the format above in different lines and if no resulted fragments just output "No related fragments")
\end{tcolorbox}

    \begin{tcolorbox}[
        colback=black!5,
        colframe=PromptPurple,
        colbacktitle=PromptPurple,
        coltitle=white,
        fonttitle=\bfseries\small,
        title={\centering Theme Identification: Staff \& Service},
        arc=2mm,
        boxrule=1.5pt,
        left=10pt, right=10pt, top=5pt, bottom=5pt, breakable
    ]
        \small
        \setlength{\parskip}{0.3em}

You are good at understanding documents with hotel review opinions.
Below is a business review for a hotel, please extract fragments that are related to Service of the hotel.

Definition of Staff \& Service:
Captures feedback on hotel personnel, focusing on their attitude and competence. This includes traits like friendliness, helpfulness, professionalism, and attentiveness. It covers all departments: front desk, check-in/out, concierge, valet, bellhop, and housekeeping. This theme also includes the speed and politeness of room service. Excludes: Feedback on the quality or taste of room service food (which belongs to the food theme) and the quality/result of the cleaning (which belongs to cleanliness).

Final extracted fragments (follow the format above in different lines and if no resulted fragments just output "No related fragments")
\end{tcolorbox}

    \begin{tcolorbox}[
        colback=black!5,
        colframe=PromptPurple,
        colbacktitle=PromptPurple,
        coltitle=white,
        fonttitle=\bfseries\small,
        title={\centering Theme Identification: Rooms},
        arc=2mm,
        boxrule=1.5pt,
        left=10pt, right=10pt, top=5pt, bottom=5pt, breakable
    ]
        \small
        \setlength{\parskip}{0.3em}

You are good at understanding documents with hotel review opinions.
Below is a business review for a hotel, please extract fragments that are related to Rooms of the hotel.
Definition of Rooms:
Captures feedback on the guest's private room, focusing on its comfort, features, and physical attributes. This includes the bed, pillows, linens, room size, furniture, and in-room decor. It also covers the functionality of in-room amenities like the TV, minibar, fridge, coffee maker, and air conditioning (AC). Opinions on the view from the room and the balcony belong here but opinions on views from public areas belong in Building \& Facilities theme. Excludes: All opinions on noise including AC noise which belong to quietness, cleanliness (which belong to cleanliness), or the bathroom (which is a separate theme).

Final extracted fragments (follow the format above in different lines and if no resulted fragments just output "No related fragments")
\end{tcolorbox}

\subsubsection{PeerSum: theme identification}
\label{subsubsec:space_theme_identification}

Theme-identification prompts for PeerSum dataset is used to compute aspect coverage in Section~\ref{subsection: eval_metrics}. We use the exact same prompt template as \cite{li2025decomposed}, except that our theme definitions are more detailed.

    \begin{tcolorbox}[
        colback=black!5,
        colframe=PromptPurple,
        colbacktitle=PromptPurple,
        coltitle=white,
        fonttitle=\bfseries\small,
        title={\centering Theme Identification: Advancement},
        arc=2mm,
        boxrule=1.5pt,
        left=10pt, right=10pt, top=5pt, bottom=5pt, breakable
    ]
        \small
        \setlength{\parskip}{0.3em}

You are good at understanding documents with scientific review opinions.
Below is a scientific review for an academic manuscript, please extract text fragments that are related to Advancement of the research work.

Definition of Advancement:
Opinions about the significance and concrete contributions of the work in advancing the research field. This includes empirical improvements over strong baselines, new capabilities, efficiency gains, theoretical or practical impact, and how much the work helps future research or applications. Advancement may be recognized even when novelty is incremental, as long as the work delivers meaningful progress. Do not use Advancement when a comment focuses only on how original or derivative the idea is; those belong to Novelty.

Final extracted fragments (follow the format above in different lines and if no resulted fragments just output "No related fragments")
\vspace{0.5em}

\textbf{Example 1} \\
        \textbf{Input:} <input\_text> \\
        \textbf{Output:} <output\_text> \\[0.5em]
\end{tcolorbox}

    \begin{tcolorbox}[
        colback=black!5,
        colframe=PromptPurple,
        colbacktitle=PromptPurple,
        coltitle=white,
        fonttitle=\bfseries\small,
        title={\centering Theme Identification: Clarity},
        arc=2mm,
        boxrule=1.5pt,
        left=10pt, right=10pt, top=5pt, bottom=5pt, breakable
    ]
        \small
        \setlength{\parskip}{0.3em}

You are good at understanding documents with scientific review opinions.
Below is a scientific review for an academic manuscript, please extract fragments that are related to Clarity of the research work

Definition of Clarity:
Opinions about the writing quality, readability, and presentation of the paper. This includes organization and flow, completeness and accessibility of explanations, quality and interpretability of figures and tables, notation and terminology, and mechanical issues such as grammar or formatting. Use Clarity for cases where ideas might be sound but are hard to follow, and for reviewer confusion or requests for clarification. Do not use Clarity when the main point is that methods, experiments, or arguments are incorrect or unsupported; those belong to Soundness.

Final extracted fragments (follow the format above in different lines and if no resulted fragments just output "No related fragments")
\vspace{0.5em}

\textbf{Example 1} \\
        \textbf{Input:} <input\_text> \\
        \textbf{Output:} <output\_text> \\[0.5em]
\end{tcolorbox}

    \begin{tcolorbox}[
        colback=black!5,
        colframe=PromptPurple,
        colbacktitle=PromptPurple,
        coltitle=white,
        fonttitle=\bfseries\small,
        title={\centering Theme Identification: Compliance},
        arc=2mm,
        boxrule=1.5pt,
        left=10pt, right=10pt, top=5pt, bottom=5pt, breakable
    ]
        \small
        \setlength{\parskip}{0.3em}

You are good at understanding documents with scientific review opinions.
Below is a scientific review for an academic manuscript, please extract fragments that are related to Compliance of the research work.

Definition of Compliance:
Opinions about the paper’s adherence to venue policies, ethical standards, and procedural guidelines. This includes fit to the venue’s scope, anonymity violations, missing ethics approvals, data/privacy or consent issues, non‑compliance in user studies, citation or formatting requirements, and other policy‑level concerns. Do not use Compliance for the scientific quality of methods or experiments (Soundness) or for the importance of the contribution (Advancement).

Final extracted fragments (follow the format above in different lines and if no resulted fragments just output "No related fragments")
\vspace{0.5em}

\textbf{Example 1} \\
        \textbf{Input:} <input\_text> \\
        \textbf{Output:} <output\_text> \\[0.5em]
\end{tcolorbox}

    \begin{tcolorbox}[
        colback=black!5,
        colframe=PromptPurple,
        colbacktitle=PromptPurple,
        coltitle=white,
        fonttitle=\bfseries\small,
        title={\centering Theme Identification: Novelty},
        arc=2mm,
        boxrule=1.5pt,
        left=10pt, right=10pt, top=5pt, bottom=5pt, breakable
    ]
        \small
        \setlength{\parskip}{0.3em}

You are good at understanding documents with scientific review opinions.
Below is a scientific meta-review for an academic manuscript, please extract fragments that are related to Novelty of the research work.

Definition of Novelty:
Opinions about the originality and innovation of the work. This includes whether the paper introduces genuinely new ideas, formulations, methods, architectures, datasets, or theoretical insights, and how distinct it is from prior or state‑of‑the‑art work. It covers phrases such as “novel contribution,” “incremental,” or “already known.” Do not use Novelty when a comment mainly evaluates the importance or practical impact of the work rather than how new it is; those belong to Advancement.

Final extracted fragments (follow the format above in different lines and if no resulted fragments just output "No related fragments")
\vspace{0.5em}

\textbf{Example 1} \\
        \textbf{Input:} <input\_text> \\
        \textbf{Output:} <output\_text> \\[0.5em]
\end{tcolorbox}

    \begin{tcolorbox}[
        colback=black!5,
        colframe=PromptPurple,
        colbacktitle=PromptPurple,
        coltitle=white,
        fonttitle=\bfseries\small,
        title={\centering Theme Identification: Soundness},
        arc=2mm,
        boxrule=1.5pt,
        left=10pt, right=10pt, top=5pt, bottom=5pt, breakable
    ]
        \small
        \setlength{\parskip}{0.3em}

You are good at understanding documents with scientific review opinions.
Below is a scientific meta-review for an academic manuscript, please extract fragments that are related to Soundness of the research work.

Definition of Soundness:
Opinions about the technical correctness, methodological validity, and strength of evidence. This includes the appropriateness of the methodology, experimental design and baselines, correctness and rigor of theory or proofs, quality of analysis, reliability of results, and whether claims are adequately supported. Use Soundness when reviewers say things like “experiments are flawed,” “evidence is convincing,” “claims are oversold,” or “proofs are not rigorous.” Do not use Soundness when the comment is mainly about readability or organization; those belong to Clarity.

Final extracted fragments (follow the format above in different lines and if no resulted fragments just output "No related fragments")
\vspace{0.5em}

\textbf{Example 1} \\
        \textbf{Input:} <input\_text> \\
        \textbf{Output:} <output\_text> \\[0.5em]
\end{tcolorbox}

    \begin{tcolorbox}[
        colback=black!5,
        colframe=PromptPurple,
        colbacktitle=PromptPurple,
        coltitle=white,
        fonttitle=\bfseries\small,
        title={\centering Theme Identification: Soundness},
        arc=2mm,
        boxrule=1.5pt,
        left=10pt, right=10pt, top=5pt, bottom=5pt, breakable
    ]
        \small
        \setlength{\parskip}{0.3em}

You are good at understanding documents with scientific review opinions.
Below is a scientific meta-review for an academic manuscript, please extract fragments that are related to Soundness of the research work.

Definition of Soundness:
Opinions about the technical correctness, methodological validity, and strength of evidence. This includes the appropriateness of the methodology, experimental design and baselines, correctness and rigor of theory or proofs, quality of analysis, reliability of results, and whether claims are adequately supported. Use Soundness when reviewers say things like “experiments are flawed,” “evidence is convincing,” “claims are oversold,” or “proofs are not rigorous.” Do not use Soundness when the comment is mainly about readability or organization; those belong to Clarity.

Final extracted fragments (follow the format above in different lines and if no resulted fragments just output "No related fragments")
\vspace{0.5em}

\textbf{Example 1} \\
        \textbf{Input:} <input\_text> \\
        \textbf{Output:} <output\_text> \\[0.5em]
\end{tcolorbox}

\end{document}